%% file: main.tex
\pdfoutput=1
\pdftrailerid{}
\documentclass[10pt,twocolumn,letterpaper]{article}
\usepackage[pagenumbers]{cvpr}
\usepackage{multirow}
\usepackage{array}
\usepackage{adjustbox}
\usepackage{algorithm}
\usepackage[noend]{algpseudocode}
\algrenewcommand\algorithmicrequire{\textbf{Input:}}
\usepackage{amsmath,amssymb}

\input{preamble}

\definecolor{cvprblue}{rgb}{0.21,0.49,0.74}
\usepackage[pagebackref,breaklinks,colorlinks,allcolors=cvprblue]{hyperref}
\newcommand{\xref}[1]{\ref{#1}}
\newcommand{\xeqref}[1]{\textup{(\xref{#1})}}

\title{Inherited Adapter Experts for Replay-Free Domain-Incremental Learning}
\author{Gyutae Oh \qquad Hyunmin Kong \qquad Jitae Shin\thanks{Corresponding author.}\\
Department of Electrical and Computer Engineering,\\
Sungkyunkwan University,\\
Suwon 16419, Republic of Korea\\
{\tt\small \{alswo740012, khm1754\}@g.skku.edu, jtshin@skku.edu}}

\begin{document}
\maketitle
\input{sec/0_abstract}
\input{sec/1_intro}
\input{sec/2_related}
\input{sec/2b_preliminaries}
\input{sec/3_method}
\input{sec/4_diagnosis}

\input{sec/5_experiments}
\input{sec/6_conclusion}

\input{sec/llm_usage}
{
    \small
    \bibliographystyle{ieeenat_fullname}
    \bibliography{main}
}
\clearpage
\renewcommand{\theHtable}{S\arabic{table}}
\renewcommand{\theHfigure}{S\arabic{figure}}
\renewcommand{\theHalgorithm}{S\arabic{algorithm}}
\input{sec/supp}

\end{document}

%% file: preamble.tex
\usepackage{tikz}
\usetikzlibrary{arrows.meta,positioning,fit}

%% file: sec/0_abstract.tex
\begin{abstract}
Replay-free domain-incremental learning trains a model on the same classes in successive domains without keeping earlier images.
Per-domain learners freeze a pre-trained ViT and train an additional expert for each new domain: the small set of parameters and the classification head learned on that domain.
Test images carry no domain label, so a router selects one expert for each image.
Recent work trains this router.
However, a training-free Mahalanobis router costs the analyzed learners under 2.7 points of final accuracy relative to always selecting each image's own-domain expert.
The experts, not this router, limit accuracy.
Our learner, Lamarck, therefore uses this router and strengthens the experts.
Its adapters, small trainable layers that correct every image token after each deep ViT block, add 4.4 points of final accuracy on the deepfake benchmark CDDB-Hard.
In addition, each expert is initialized from its predecessor at no extra cost: the domains share their classes, so an image misrouted to a later expert loses less.
Against twelve frozen-backbone baselines under one protocol, Lamarck is first on CDDB-Hard by 5.2 points, second on DomainNet and OfficeHome, and highest in mean accuracy.
\end{abstract}

%% file: sec/1_intro.tex
\section{Introduction}
\label{sec:intro}
Domain-incremental learning (DIL) trains one model on a fixed set of classes in successive stages.
Each stage brings a new kind of image, called a \emph{domain}~\cite{van2022three,wang2024comprehensive}.
For example, a bicycle is learned from clipart, then doodles, then photographs, and a deepfake detector learns each new source of fake images~\cite{li2023cddb}.
We study the replay-free setting, in which no image of an earlier domain is kept; only model parameters and feature statistics remain.
Training on a new domain then lowers accuracy on the earlier ones (\emph{forgetting}), and a test image does not say which domain it comes from.

\begin{figure}[t]
\centering
\includegraphics[width=0.96\linewidth]{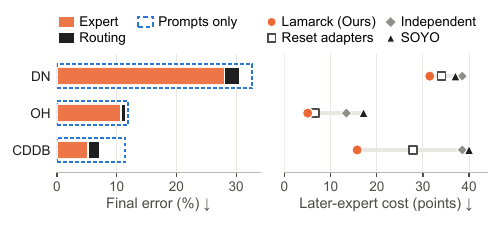}
\caption{\textbf{Most of the final error is expert error (left); inheritance lowers the accuracy lost at later experts (right).}
Left: Lamarck's final error (100$-$final accuracy) on DomainNet (DN), OfficeHome (OH) and CDDB-Hard (CDDB), split into the error of each image's own-domain expert (Expert) and the error added by the Mahalanobis router (Routing).
Dashed (Prompts only): Lamarck without adapters.
Right: later-expert cost (Eq.~\eqref{eq:costs}), the accuracy a domain's images lose at an expert trained after their own domain.
Shown for Lamarck, Lamarck with only the adapters re-initialized for every domain (Reset adapters), Lamarck trained as independent experts (Independent) and SOYO~\cite{wang2025soyo}.
Five-seed test means.}
\label{fig:motivation}
\end{figure}

Prompt-based continual learning (PCL) freezes a pre-trained ViT and trains a classification head and \emph{prompts}.
Prompts are a few learnable vectors that the frozen blocks attend to as extra tokens~\cite{wang2022learning,wang2022dualprompt}.
Many prompt learners retrieve prompts for each image from a \emph{prompt pool} by matching the image's feature to learned keys~\cite{wang2022learning,smith2023coda,kim2024one}.
Prompt learners differ in what they share across domains.
\emph{Shared-pool learners} train one pool and one head on every domain in turn, so forgetting can remain: on DomainNet, CODA-Prompt~\cite{smith2023coda} averages 71.3\% right after learning a domain but keeps 50.6\% after the last (Sec.~\xref{sec:supp_sharedpool}).

To avoid this forgetting, \emph{per-domain learners} train an additional \emph{expert}~\cite{aljundi2017expert,byun2026expert} for each new domain: the small set of parameters and the head learned on that domain.
Each expert is frozen and kept, so later training cannot lower its accuracy.
An S-iPrompt~\cite{wang2022sprompts} expert has one prompt and one head, and a SOYO~\cite{wang2025soyo} expert adds \emph{adapters}, small trainable layers after all twelve blocks.
Since the domain of a test image is unknown, a \emph{router} (a domain identifier or selector in prior work) selects one expert for each image, and only that expert classifies it.
Such a learner can thus make two kinds of error: the router selects another expert, or the image's own-domain expert misclassifies it.
Subsequent per-domain learners target the first kind of error by training the router~\cite{wang2023hierarchical,wang2025soyo,paeedeh2025conec}.
Meanwhile, the nine prompt learners we compare place prompts at the input or in the early blocks (1--5) and train nothing in the deep blocks (6--12).

We run every expert of S-iPrompt and SOYO on every test image (Sec.~\ref{sec:diagnosis}).
A training-free router built from feature statistics costs them under 2.7 points of final accuracy (ACC: after the last domain, averaged over domains) relative to always selecting each image's own-domain expert.
It also gives both learners higher ACC than their own routers on all three benchmarks (Tabs.~\ref{tab:router} and~\ref{tab:system}).
This router picks the domain whose mean of frozen-ViT features is nearest in Mahalanobis distance~\cite{lee2018simple}.
It misroutes 3.6 to 16.2\% of the test images, but the expert it selects predicts the same classes and is often still correct.
The cost of routing is thus roughly the misrouting rate times the accuracy an image loses at the wrong expert.
The experts, not this router, limit accuracy, even for our learner, Lamarck (Fig.~\ref{fig:motivation}, left).
Under this router on DomainNet, SOYO's experts score 3.4 ACC points higher than S-iPrompt's, more than a perfect router would add to either learner.
The gap is uneven: in own-domain accuracy, the experts are nearly equal on real photographs but far apart on QuickDraw doodles (Tab.~\xref{tab:supp_perdomain}).

Lamarck, a per-domain learner, therefore uses this router and strengthens the experts.
Each expert retrieves prompts from its own prompt pool in the early blocks, as prompt learners do.
After each deep block, an adapter~\cite{houlsby2019parameter,chen2022adaptformer} adds a learned correction to every image token.
Against the same experts without adapters (Lamarck's prompt-only experts), the adapters leave own-domain accuracy on DomainNet's real photographs unchanged but raise it by 5.7 points on QuickDraw doodles.
They raise ACC on all three benchmarks, from 0.3 points on OfficeHome to 4.4 on the deepfake benchmark CDDB-Hard (Tabs.~\ref{tab:router} and~\ref{tab:modules}).
On CDDB-Hard, deep-block prompts with as many parameters as the adapters recover most of this gain, and the adapter form adds the rest (Sec.~\ref{sec:gain}).

No earlier image is kept, but the previous expert is.
Because all domains share the classes, this expert is already a classifier for the next domain.
Lamarck therefore initializes each expert, adapters included, from its predecessor, as HiDe-Prompt~\cite{wang2023hierarchical} does for its prompts.
This \emph{inheritance} (hence the name) adds no parameters or computation.
Experts trained later are then more accurate on earlier domains than independently trained ones (82.1\% against 59.4\% on CDDB-Hard; chance: 50\%), so an image misrouted to a later expert loses less (Fig.~\ref{fig:motivation}, right).
On CDDB-Hard, the inherited adapters account for about half of this advantage (Sec.~\ref{sec:inherit}).
Inheritance matters more as more images reach later experts.
Under the Mahalanobis router, which misroutes few such images, inherited experts lead independent ones by at most 0.6 ACC points.
With a weaker router fitted on two images per domain, they lead by 1.7 to 4.4.

Lamarck thus combines a training-free router, adapter experts and inheritance (Fig.~\ref{fig:method}).
Like Lamarck, SOYO keeps adapter experts, but it trains its router and independent experts.
We compare Lamarck with twelve replay-free learners trained under one protocol on the same frozen ViT, over five seeds on DomainNet (DN), OfficeHome (OH) and CDDB-Hard (CDDB).
Lamarck leads on CDDB by 5.2 ACC points, is second on DN and OH, and has the highest mean ACC, 83.6 against 79.7 for HiDe-Prompt (Tab.~\ref{tab:system}).
Inference runs the ViT twice per image, once to route and once to classify, as in other two-pass learners (Fig.~\ref{fig:efficiency}).

Our contributions are:
\begin{itemize}
\item We show on three benchmarks that a training-free Mahalanobis router costs the per-domain learners S-iPrompt, SOYO and Lamarck under 2.7 ACC points. On each benchmark, this cost is less than the largest ACC difference between their experts under that router.
\item We propose Lamarck, whose experts have prompts in the early blocks and adapters on every image token after each deep block, and test their form, tokens and placement.
\item We show that inheritance, adapters included, adds no parameters or computation. It makes later experts more accurate on earlier domains, so its gain grows as a router misroutes more images to them.
\end{itemize}\unskip %

%% file: sec/2_related.tex
\section{Related Work}
\label{sec:related}
\paragraph{Prompt-based continual learning.}
L2P~\cite{wang2022learning} queries a shared pool of input prompts with the image's feature from a separate, unprompted pass of the frozen ViT.
DualPrompt~\cite{wang2022dualprompt} places prompts as attention prefixes in blocks chosen by search: a shared prompt in blocks 1--2 and a per-task prompt in blocks 3--5.
Subsequent learners prompt the same blocks, mostly with prefixes~\cite{smith2023coda,kim2024one,xu2025kaprompt,lim2026critical,oh2025towards}.
CODA-Prompt, KA-Prompt and CPS-Prompt weight or select prompts with the unprompted-pass query, whereas OS-Prompt, OS-Prompt++ and UniPrompt-CL use the class token of the prompted pass.
None trains parameters in the deep blocks.
Lamarck's experts use OS-Prompt's query, but in its last prompt-only expert, retrieval is nearly input-independent (Sec.~\ref{sec:gain}), in line with single-prompt results in class-incremental~\cite{huang2024ovor} and online task-free~\cite{park2026prompt} settings.

\paragraph{Expert retention and routing.}
S-iPrompt~\cite{wang2022sprompts} routes its input-prompt experts by the nearest k-means centre of frozen features.
SOYO~\cite{wang2025soyo} pairs an input prompt with adapters after all twelve blocks and trains its router on features resampled from a Gaussian mixture.
G2D~\cite{byun2026expert} trains its router on generated data.
Others propose different experts or routers~\cite{wistuba2023color,wang2024nonexemplar,tang2024mind}.
In Hash~\cite{li2026hash}, the ``router'' gates a shared prompt pool per task.
Like many routers~\cite{wang2022sprompts,tang2024mind,paeedeh2026fewshot}, Lamarck's router uses statistics of frozen features in a shrunk Mahalanobis score~\cite{lee2018simple,goswami2023fecam}.
Prior work separates within-task and task-identity prediction~\cite{kim2022theoretical,wang2023hierarchical}, reports domain-identification accuracy and accuracy with domain labels~\cite{wang2022sprompts,wang2025soyo,paeedeh2025conec}, or evaluates every prompt with every head~\cite{anonymous2026routing}.
Because all DIL experts predict the same classes, a misrouted image can still be classified correctly.
We therefore measure and lower the accuracy lost per misroute.

\paragraph{Adapters and feature adaptation.}
Prompts~\cite{jia2022visual} add tokens to a frozen network's input, whereas adapters and low-rank updates change the transformation it applies: residual bottleneck adapters~\cite{houlsby2019parameter}, their parallel form for ViTs in AdaptFormer~\cite{chen2022adaptformer}, and LoRA~\cite{hu2022lora}.
Prefix tuning can be written as an adapter on the attention output~\cite{he2022towards}, but cannot change the relative attention among image tokens in its layer~\cite{petrov2024when}.
Adapter-based continual learners combine or select per-task adapters~\cite{zhou2024expandable,sun2025mos} or adapt a single model~\cite{tan2024semantically,zhou2024revisiting,mcdonnell2023ranpac}, mostly for a growing label space.
For DIL, CONEC-LoRA~\cite{paeedeh2025conec} keeps per-domain low-rank updates in blocks 7--12 and trains its router.
DUCT~\cite{zhou2025dual} instead consolidates a fine-tuned backbone and its head across domains.
Among our baselines, SOYO, EASE~\cite{zhou2024expandable} and CONEC-LoRA train per-domain parameters in the deep blocks, like Lamarck.
Adapters in all blocks (1--12) change Lamarck's ACC by at most 0.3 (Tabs.~\ref{tab:variants} and~\xref{tab:supp_rank}).

\paragraph{Function-preserving initialization.}
Expert Gate~\cite{aljundi2017expert} initializes a new expert from the most related previous expert, and HiDe-Prompt~\cite{wang2023hierarchical} initializes each task's prompt from the previous one.
Net2Net~\cite{chen2016net2net} initializes a larger model with exactly the function of a smaller one.
Likewise, LoRA zero-initializes one of its two factors~\cite{hu2022lora}, so training starts from the pre-trained function.
Lamarck applies both ideas to whole experts, and Sec.~\ref{sec:inherit} measures the effect.

%% file: sec/2b_preliminaries.tex
\section{Preliminaries}
\label{sec:prelim}
\begin{figure*}[t]
\centering
\includegraphics[width=\textwidth]{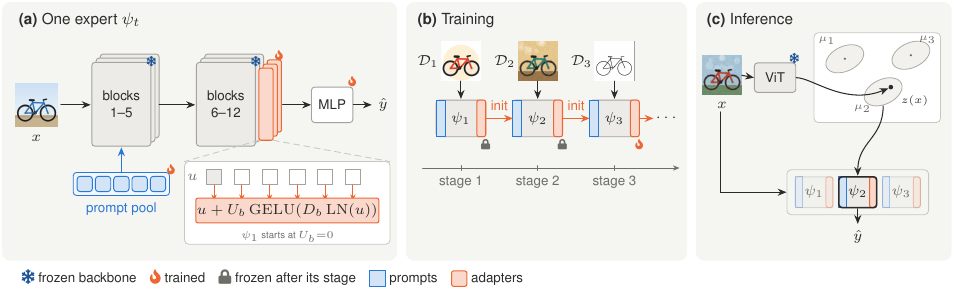}
\caption{\textbf{Lamarck.} (a) One expert on the frozen backbone. Prompts retrieved from the expert's prompt pool enter the early blocks, and after each deep block a residual adapter corrects every image token. An MLP readout on the final CLS feature computes the logits. (b) Expert $t$ is initialized from expert $t{-}1$, trained on domain $t$ and then frozen. (c) A training-free router selects the domain whose mean feature is nearest to the frozen-ViT feature $z(x)$ in shrunk Mahalanobis distance, and only that domain's expert classifies $x$.}
\label{fig:method}
\end{figure*}
\paragraph{Replay-free domain-incremental learning.}
A learner receives the training sets $\mathcal{D}_1,\dots,\mathcal{D}_T$ of $T$ domains in sequence.
Each $\mathcal{D}_t$ is drawn from $\mathcal{P}_t(x,y)$ over images $x$ and labels $y\in\mathcal{Y}$.
All domains share the label set $\mathcal{Y}$ and differ in their input distributions $\mathcal{P}_t(x)$.
At stage $t$ only $\mathcal{D}_t$ is available: from earlier stages, the learner may keep model parameters and feature statistics but no image.
Stage boundaries are known during training, but test images carry no domain label.
After every stage, the learner must classify images of all seen domains.

\paragraph{Experts and routers.}
An expert $\psi_e$ consists of the parameters and classification head trained on domain $e$ and then frozen.
With the frozen ViT backbone $\Theta$ ($M{=}12$ blocks), $\psi_e$ is a complete classifier over $\mathcal{Y}$.
A per-domain learner keeps its experts in an \emph{expert bank} and, after stage $t$, predicts
\begin{equation}
\hat{y}(x)=h_{\rho_t(x)}(x),\qquad \rho_t(x)\in\{1,\dots,t\},
\label{eq:bank}
\end{equation}
where the router $\rho_t$ selects one expert per image and $h_e(x)$ is the class predicted by expert $e$.

\paragraph{Prompts and prompt-only experts.}
A prompt is a few learnable vectors that a frozen block attends to as extra tokens, in the token sequence~\cite{jia2022visual} or as a prefix in its attention~\cite{wang2022dualprompt}.
A prompt pool holds key-prompt pairs as rows.
\emph{Prompt retrieval} weights the rows by matching a feature of the image, the \emph{query}, to the keys.
A prompt-only expert has only prompts and a head.
In Lamarck's prompt-only expert, the first $N$ blocks receive prompts, retrieved with the query of OS-Prompt~\cite{kim2024one}.
At block $i\le N$, the CLS token $Q_i$ of the incoming sequence $f_i$ queries a pool of $Z$ rows of keys and one-token prompts, $K_i,P_i\in\mathbb{R}^{Z\times 768}$.
Each stage makes $\lfloor Z/T\rfloor$ new rows visible, plus the remainder at the last stage.
Retrieval at stage $t$ uses all $Z_t$ visible rows, including the frozen rows of earlier stages.
With cosine similarity $\gamma$, the retrieved prompt
\begin{equation}
\phi_i=\sum_{k=1}^{Z_t}\frac{\exp\gamma(Q_i,K_{i,k})}{\sum_{k'}\exp\gamma(Q_i,K_{i,k'})}\,P_{i,k}
\label{eq:retrieval}
\end{equation}
is prepended to the sequence, $f_{i+1}=\Theta_i([\phi_i,f_i])$, and stays there, as in UniPrompt-CL~\cite{oh2025towards}.
Through attention, the prompt tokens change which features the frozen blocks combine, not the transformation each block applies.

\paragraph{Metrics.}
Let $A_{t,j}$ be the accuracy on the test split (TEST) of domain $j$ after stage $t$.
We report ACC and the average incremental accuracy (AIA),
\begin{equation}
\mathrm{ACC}=\frac{1}{T}\sum_{j=1}^{T}A_{T,j},\qquad
\mathrm{AIA}=\frac{1}{T}\sum_{t=1}^{T}\frac{1}{t}\sum_{j=1}^{t}A_{t,j},
\label{eq:metrics}
\end{equation}
which weight domains equally, as in KA-Prompt~\cite{xu2025kaprompt}.
To separate the experts' errors from the router's, let $a_{e,j}$ be the accuracy of expert $e$ on domain $j$ after the last stage, and let $C(x)$ be the set of experts that classify image $x$ correctly:
\begin{equation}
\begin{aligned}
\mathrm{DRA}&=\tfrac{1}{T}\textstyle\sum_{j}a_{j,j},\quad \text{routing loss}=\mathrm{DRA}-\mathrm{ACC},\\
\mathrm{SE}&=\tfrac{1}{T}\textstyle\sum_{j}\Pr_{x\in\mathrm{TEST}_j}\big[C(x)=\emptyset\big].
\end{aligned}
\label{eq:diag}
\end{equation}
The domain-routed accuracy (DRA) is the accuracy when the domain label selects each image's own-domain expert.
The shared error (SE) is the share of images that no router can recover.
The quantities below evaluate images of domain $j$ at a later expert $e{>}j$ or an earlier expert $e{<}j$.
Each cost is the accuracy lost relative to the own-domain expert:
\begin{equation}
\begin{aligned}
\text{later-expert accuracy}&=\tfrac{2}{T(T-1)}\textstyle\sum_{e>j}a_{e,j},\\
\text{later-expert cost}&=\tfrac{2}{T(T-1)}\textstyle\sum_{e>j}\big(a_{j,j}-a_{e,j}\big),\\
\text{earlier-expert cost}&=\tfrac{2}{T(T-1)}\textstyle\sum_{e<j}\big(a_{j,j}-a_{e,j}\big).
\end{aligned}
\label{eq:costs}
\end{equation}

%% file: sec/3_method.tex
\section{Lamarck}
\label{sec:method}
\paragraph{Prompt base.}
Each Lamarck expert $\psi_e$ (Fig.~\ref{fig:method}a) adds adapters to the prompt-only expert of Sec.~\ref{sec:prelim}, whose head is an MLP readout.
As in most compared prompt learners, prompts enter the $N{=}5$ early blocks~\cite{kim2024one,oh2025towards}, each with a pool of $Z{=}100$ rows~\cite{smith2023coda,kim2024one}.

\paragraph{Deep adapters.}
After each deep block $b\in\mathcal{B}=\{N{+}1,\dots,M\}$, a residual bottleneck adapter~\cite{houlsby2019parameter} corrects every image token $u$ (the CLS and 196 patch tokens, not prompt tokens):
\begin{equation}
u\leftarrow u+U_b\,\mathrm{GELU}\big(D_b\,\mathrm{LN}(u)\big),
\label{eq:adapter}
\end{equation}
with down- and up-projections $D_b\in\mathbb{R}^{r\times d}$, $U_b\in\mathbb{R}^{d\times r}$ ($d{=}768$, rank $r{=}32$), no biases and parameter-free layer normalization.
Correcting only the CLS token is less accurate (Tab.~\ref{tab:variants}).
Unlike prompts, the adapters change the features themselves.
They add $2rd|\mathcal{B}|\approx0.34$M parameters per expert, against 86M in the frozen backbone.

\paragraph{Readout and training.}
The readout is a two-layer MLP $f_{\mathrm{fc}}$ with hidden width 384 and ReLU.
It maps $\hat f$, the CLS feature after the final layer normalization, to logits scaled by the sigmoid $\sigma$ of a learned scalar $w$:
\begin{equation}
\mathrm{logits}(x)=\sigma(w)\,f_{\mathrm{fc}}(\hat f),\qquad \mathcal{L}=\mathrm{CE}(\mathrm{logits}(x),y).
\label{eq:readout}
\end{equation}
At stage $t$, Lamarck trains the new prompt rows, all adapters and the readout on $\mathcal{D}_t$, then freezes and stores expert $\psi_t$.

\paragraph{Inheritance.}
Lamarck initializes expert $t$ with the prompt rows, adapters and readout of its predecessor and adds only new prompt rows (Fig.~\ref{fig:method}b).
The first expert's up-projections $U_b$ start at zero, so it begins as a prompt-only expert.
Inheritance needs only the stored predecessor, not earlier images, and adds no parameters or computation.
Because each expert starts from its predecessor, an image misrouted to a later expert loses less accuracy than with independently trained experts (Fig.~\ref{fig:heatmap}; Sec.~\ref{sec:inherit}).

\input{sec/table_router}
\paragraph{Mahalanobis router.}
An unprompted pass gives the $\ell_2$-normalized CLS feature $z(x)$.
For every seen domain $e$, the router keeps the mean $\mu_e$ and the covariance of $z$ over the training split (TRAIN), and the number of images.
It pools the covariances into the within-domain covariance $S$ and shrinks $S$ toward a scaled identity, $\Sigma=(1-\lambda)S+\lambda\,\tfrac{\mathrm{tr}\,S}{d}I$, with $\lambda{=}0.01$ chosen on the validation split (VAL).
The router selects the domain whose mean is nearest to $z(x)$ in shrunk Mahalanobis distance~\cite{lee2018simple,goswami2023fecam}:
\begin{equation}
\rho_t(x)=\underset{e\le t}{\arg\min}\;\big(z(x)-\mu_e\big)^{\top}\Sigma^{-1}\big(z(x)-\mu_e\big).
\label{eq:router}
\end{equation}
It keeps only these feature statistics and needs no training.
Inference costs one unprompted and one prompted pass (Alg.~\xref{alg:lamarck}).

%% file: sec/table_router.tex
\begin{table*}[t]
\centering
\caption{\textbf{Under the Mahalanobis router, the routing loss is smaller than the ACC spread across expert banks.} ACC of each expert bank under two routers on the same frozen features: Mahalanobis (Mah., Eq.~\eqref{eq:router}) and prototype (Proto.; cosine similarity to eight Gaussian-mixture means per domain). DRA (Eq.~\eqref{eq:diag}): domain labels select own-domain experts. Routing loss$\,{=}\,$DRA$-$Mah. SE: share of images that every stored expert misclassifies. TEST, five-seed mean$\pm$s.d. Tab.~\ref{tab:system} reports S-iPrompt and SOYO with their own routers.}
\label{tab:router}
\scriptsize
\setlength{\tabcolsep}{2.4pt}
\resizebox{\textwidth}{!}{%
\begin{tabular}{@{}lcccccccccccc@{}}
\toprule
 & \multicolumn{4}{c}{DN} & \multicolumn{4}{c}{OH} & \multicolumn{4}{c}{CDDB} \\
\cmidrule(lr){2-5}\cmidrule(lr){6-9}\cmidrule(l){10-13}
Expert bank & DRA$\uparrow$ & Mah.$\uparrow$ & Proto.$\uparrow$ & SE$\downarrow$ & DRA$\uparrow$ & Mah.$\uparrow$ & Proto.$\uparrow$ & SE$\downarrow$ & DRA$\uparrow$ & Mah.$\uparrow$ & Proto.$\uparrow$ & SE$\downarrow$ \\
\midrule
S-iPrompt & $69.26{\scriptstyle\pm0.08}$ & $66.79{\scriptstyle\pm0.08}$ & $64.95{\scriptstyle\pm0.14}$ & $23.43{\scriptstyle\pm0.10}$ & $88.81{\scriptstyle\pm0.23}$ & $87.89{\scriptstyle\pm0.19}$ & $86.64{\scriptstyle\pm0.38}$ & $6.28{\scriptstyle\pm0.22}$ & $86.99{\scriptstyle\pm2.65}$ & $87.12{\scriptstyle\pm1.97}$ & $84.54{\scriptstyle\pm1.15}$ & $1.27{\scriptstyle\pm1.09}$ \\
SOYO & $72.81{\scriptstyle\pm0.07}$ & $70.19{\scriptstyle\pm0.08}$ & $68.31{\scriptstyle\pm0.17}$ & $19.96{\scriptstyle\pm0.06}$ & $87.51{\scriptstyle\pm0.50}$ & $86.36{\scriptstyle\pm0.46}$ & $84.61{\scriptstyle\pm0.62}$ & $6.50{\scriptstyle\pm0.34}$ & $88.15{\scriptstyle\pm1.23}$ & $86.46{\scriptstyle\pm1.35}$ & $84.17{\scriptstyle\pm1.29}$ & $3.64{\scriptstyle\pm1.83}$ \\
Lamarck, prompts only\hspace{0.2em} & $69.72{\scriptstyle\pm0.16}$ & $67.32{\scriptstyle\pm0.16}$ & $65.62{\scriptstyle\pm0.32}$ & $22.90{\scriptstyle\pm0.11}$ & $88.91{\scriptstyle\pm0.33}$ & $88.14{\scriptstyle\pm0.22}$ & $87.21{\scriptstyle\pm0.14}$ & $6.91{\scriptstyle\pm0.11}$ & $89.35{\scriptstyle\pm1.18}$ & $88.57{\scriptstyle\pm0.92}$ & $87.10{\scriptstyle\pm1.32}$ & $3.27{\scriptstyle\pm0.51}$ \\
\textbf{Lamarck (Ours)} & $71.93{\scriptstyle\pm0.12}$ & $69.46{\scriptstyle\pm0.14}$ & $67.75{\scriptstyle\pm0.13}$ & $20.59{\scriptstyle\pm0.13}$ & $89.29{\scriptstyle\pm0.24}$ & $88.47{\scriptstyle\pm0.25}$ & $87.43{\scriptstyle\pm0.25}$ & $6.22{\scriptstyle\pm0.30}$ & $94.78{\scriptstyle\pm0.53}$ & $92.94{\scriptstyle\pm0.62}$ & $91.61{\scriptstyle\pm0.73}$ & $1.80{\scriptstyle\pm0.22}$ \\
\bottomrule
\end{tabular}}
\end{table*}

%% file: sec/4_diagnosis.tex
\section{The Experts Limit Accuracy}
\label{sec:diagnosis}
To separate routing loss from expert error, we analyze four expert banks (Sec.~\ref{sec:setup}): S-iPrompt's prompt-only experts, SOYO's adapter experts, and Lamarck's experts with and without adapters.

\paragraph{Evaluation without retraining.}
We evaluate every stored expert on the test images of every seen domain.
With each learner's own router, this evaluation reproduces its routed predictions exactly (Sec.~\xref{sec:supp_settings}).
This extends SOYO's evaluation of own-domain experts~\cite{wang2025soyo} to every expert--domain pair.
It also extends the evaluation of every prompt with every head in~\cite{anonymous2026routing} to whole experts, including adapter experts.
We route all banks with two training-free routers on frozen features: the Mahalanobis router (Eq.~\eqref{eq:router}) and a prototype router.
The prototype router fits eight Gaussian-mixture means per domain on TRAIN and selects the domain of the mean most cosine-similar to the image's frozen feature.
All banks thus receive the same routes and differ only in their experts.
A misrouted image can be classified correctly where its own-domain expert errs, so ACC can exceed DRA (Eq.~\eqref{eq:diag}), as for S-iPrompt on CDDB under the Mahalanobis router.

\paragraph{The routing loss is small.}
Under the Mahalanobis router, the routing loss is at most 2.62 points on DN, 1.15 on OH and 1.84 on CDDB (Tab.~\ref{tab:router}).
The expert error (100$-$DRA) is larger: 11.2 to 30.7 for S-iPrompt and SOYO and 5.2 to 28.1 for full Lamarck (Tab.~\ref{tab:router}).
The router misroutes 11.9\%, 16.2\% and 3.6\% of TEST images (Tab.~\xref{tab:supp_inherit_routing}), but the expert it selects is often still correct.
Replacing the experts under this router can change ACC by more than the routing loss: even the three banks without Lamarck's adapters span 3.40 points on DN (SOYO over S-iPrompt), 1.78 on OH and 2.11 on CDDB (Tab.~\ref{tab:router}).
The prototype router ranks the banks identically.

\input{sec/table1_accuracy}

%% file: sec/table1_accuracy.tex
\begin{table*}[t]
\centering
\caption{\textbf{Lamarck has the best ACC among frozen-backbone learners on CDDB and averaged over the three benchmarks, and the second best on DN and OH.} Five-seed TEST accuracy (\%, mean$\pm$s.d.). For a fair comparison, all methods are trained under one protocol (Sec.~\ref{sec:setup}), each with its own inference rule. Bold, underlined: best and second best. The joint linear probe, trained on all domains at once, is a reference outside this comparison. SLCA~\cite{zhang2023slca} and DUCT~\cite{zhou2025dual}, which fine-tune the backbone, are in Tab.~\xref{tab:supp_duct}.}
\label{tab:system}
\scriptsize
\setlength{\tabcolsep}{3pt}
\renewcommand{\arraystretch}{1.0}
\begin{tabular}{@{}llcccccccc@{}}
\toprule
\multirow{2}{*}{Method} & \multirow{2}{*}{Venue} & \multicolumn{2}{c}{DN} & \multicolumn{2}{c}{OH} & \multicolumn{2}{c}{CDDB} & \multicolumn{2}{c}{Mean} \\
\cmidrule(lr){3-4}\cmidrule(lr){5-6}\cmidrule(lr){7-8}\cmidrule(l){9-10}
& & AIA$\uparrow$ & ACC$\uparrow$ & AIA$\uparrow$ & ACC$\uparrow$ & AIA$\uparrow$ & ACC$\uparrow$ & AIA$\uparrow$ & ACC$\uparrow$ \\
\midrule
Joint linear probe & -- & -- & $64.08\,{\scriptstyle\pm\,0.04}$ & -- & $86.95\,{\scriptstyle\pm\,0.37}$ & -- & $80.63\,{\scriptstyle\pm\,0.34}$ & -- & $77.22$ \\
\midrule
\multicolumn{10}{@{}l}{\textit{Shared-pool learners: one prompt pool and classifier shared by all domains}} \\
CODA-Prompt~\cite{smith2023coda} & CVPR 2023 & $54.55\,{\scriptstyle\pm\,0.10}$ & $50.62\,{\scriptstyle\pm\,0.14}$ & $84.46\,{\scriptstyle\pm\,0.21}$ & $86.62\,{\scriptstyle\pm\,0.23}$ & $87.04\,{\scriptstyle\pm\,0.58}$ & $72.92\,{\scriptstyle\pm\,2.33}$ & $75.35$ & $70.05$ \\
OS-Prompt~\cite{kim2024one} & ECCV 2024 & $52.60\,{\scriptstyle\pm\,0.12}$ & $49.74\,{\scriptstyle\pm\,0.19}$ & $84.22\,{\scriptstyle\pm\,0.27}$ & $86.31\,{\scriptstyle\pm\,0.51}$ & $86.49\,{\scriptstyle\pm\,0.43}$ & $72.84\,{\scriptstyle\pm\,2.14}$ & $74.44$ & $69.63$ \\
OS-Prompt++~\cite{kim2024one} & ECCV 2024 & $52.55\,{\scriptstyle\pm\,0.07}$ & $49.74\,{\scriptstyle\pm\,0.23}$ & $84.26\,{\scriptstyle\pm\,0.31}$ & $86.33\,{\scriptstyle\pm\,0.51}$ & $86.53\,{\scriptstyle\pm\,0.39}$ & $72.69\,{\scriptstyle\pm\,2.21}$ & $74.45$ & $69.59$ \\
KA-Prompt~\cite{xu2025kaprompt} & ICML 2025 & $49.84\,{\scriptstyle\pm\,0.03}$ & $50.60\,{\scriptstyle\pm\,0.04}$ & $83.73\,{\scriptstyle\pm\,0.47}$ & $85.90\,{\scriptstyle\pm\,0.22}$ & $81.84\,{\scriptstyle\pm\,0.33}$ & $74.96\,{\scriptstyle\pm\,0.39}$ & $71.80$ & $70.49$ \\
CPS-Prompt~\cite{lim2026critical} & CVPR 2026 & $53.64\,{\scriptstyle\pm\,0.10}$ & $49.97\,{\scriptstyle\pm\,0.14}$ & $83.96\,{\scriptstyle\pm\,0.35}$ & $85.84\,{\scriptstyle\pm\,0.42}$ & $85.75\,{\scriptstyle\pm\,0.71}$ & $71.45\,{\scriptstyle\pm\,1.47}$ & $74.45$ & $69.09$ \\
UniPrompt-CL~\cite{oh2025towards} & WACV 2027 & $51.77\,{\scriptstyle\pm\,0.20}$ & $47.55\,{\scriptstyle\pm\,0.29}$ & $85.61\,{\scriptstyle\pm\,0.27}$ & $87.03\,{\scriptstyle\pm\,0.34}$ & $89.42\,{\scriptstyle\pm\,0.51}$ & $76.14\,{\scriptstyle\pm\,0.89}$ & $75.60$ & $70.24$ \\
\midrule
\multicolumn{10}{@{}l}{\textit{Per-domain learners: parameters trained and kept for each domain}} \\
S-iPrompt~\cite{wang2022sprompts} & NeurIPS 2022 & $65.51\,{\scriptstyle\pm\,0.10}$ & $65.81\,{\scriptstyle\pm\,0.03}$ & $84.71\,{\scriptstyle\pm\,0.37}$ & $86.99\,{\scriptstyle\pm\,0.40}$ & $90.23\,{\scriptstyle\pm\,0.81}$ & $85.60\,{\scriptstyle\pm\,1.92}$ & $80.15$ & $79.47$ \\
HiDe-Prompt~\cite{wang2023hierarchical} & NeurIPS 2023 & $64.54\,{\scriptstyle\pm\,0.08}$ & $64.51\,{\scriptstyle\pm\,0.08}$ & $85.92\,{\scriptstyle\pm\,0.36}$ & $88.12\,{\scriptstyle\pm\,0.21}$ & $91.14\,{\scriptstyle\pm\,0.34}$ & $86.39\,{\scriptstyle\pm\,0.61}$ & $80.53$ & $\underline{79.67}$ \\
EASE~\cite{zhou2024expandable} & CVPR 2024 & $62.09\,{\scriptstyle\pm\,0.03}$ & $61.36\,{\scriptstyle\pm\,0.02}$ & $85.00\,{\scriptstyle\pm\,0.10}$ & $86.66\,{\scriptstyle\pm\,0.08}$ & $79.79\,{\scriptstyle\pm\,0.77}$ & $67.14\,{\scriptstyle\pm\,0.79}$ & $75.63$ & $71.72$ \\
SOYO~\cite{wang2025soyo} & CVPR 2025 & $\mathbf{69.33}\,{\scriptstyle\pm\,0.08}$ & $\mathbf{69.91}\,{\scriptstyle\pm\,0.08}$ & $82.91\,{\scriptstyle\pm\,0.85}$ & $85.05\,{\scriptstyle\pm\,0.60}$ & $89.96\,{\scriptstyle\pm\,0.49}$ & $79.24\,{\scriptstyle\pm\,0.95}$ & $80.73$ & $78.07$ \\
Hash~\cite{li2026hash} & CVPR 2026 & $62.42\,{\scriptstyle\pm\,0.61}$ & $58.89\,{\scriptstyle\pm\,0.70}$ & $\mathbf{86.31}\,{\scriptstyle\pm\,0.20}$ & $\mathbf{88.64}\,{\scriptstyle\pm\,0.12}$ & $\underline{92.37}\,{\scriptstyle\pm\,0.36}$ & $\underline{87.75}\,{\scriptstyle\pm\,0.88}$ & $80.37$ & $78.43$ \\
CONEC-LoRA~\cite{paeedeh2025conec} & arXiv 2026 & $67.97\,{\scriptstyle\pm\,0.05}$ & $68.15\,{\scriptstyle\pm\,0.13}$ & $\underline{86.28}\,{\scriptstyle\pm\,0.23}$ & $87.80\,{\scriptstyle\pm\,0.16}$ & $90.31\,{\scriptstyle\pm\,1.45}$ & $79.96\,{\scriptstyle\pm\,2.34}$ & $\underline{81.52}$ & $78.64$ \\
\midrule
\textbf{Lamarck (Ours)} & -- & $\underline{68.89}\,{\scriptstyle\pm\,0.20}$ & $\underline{69.46}\,{\scriptstyle\pm\,0.14}$ & $85.85\,{\scriptstyle\pm\,0.63}$ & $\underline{88.47}\,{\scriptstyle\pm\,0.25}$ & $\mathbf{97.16}\,{\scriptstyle\pm\,0.17}$ & $\mathbf{92.94}\,{\scriptstyle\pm\,0.62}$ & $\mathbf{83.97}$ & $\mathbf{83.62}$ \\
\bottomrule
\end{tabular}
\end{table*}

%% file: sec/5_experiments.tex
\section{Experiments}
\label{sec:experiments}

\subsection{Setup}
\label{sec:setup}
\paragraph{Benchmarks.}
As in prior work, we use DN~\cite{peng2019moment} (six domains from clipart to sketch, 345 classes), OH~\cite{venkateswara2017deep} (four domains, 65 classes) and CDDB~\cite{li2023cddb} (five domains from GauGAN to SAN, real vs.\ fake).
DUCT~\cite{zhou2025dual} and CONEC-LoRA~\cite{paeedeh2025conec} evaluate on all three, and S-Prompts~\cite{wang2022sprompts} and SOYO~\cite{wang2025soyo} evaluate on DN and CDDB.
Each benchmark's domains are learned in a fixed order.
DN keeps its official test set as TEST, and we split OH and CDDB into TRAIN, VAL and TEST (Sec.~\xref{sec:supp_impl}).
CDDB's domains are unbalanced: its two smallest, WhichFaceIsReal (WFIR) and SAN, contain 160 and 33 of its 2{,}140 TEST images.

\paragraph{Baselines.}
The twelve baselines (Tab.~\ref{tab:system}) are six shared-pool learners and six per-domain learners.
We exclude G2D~\cite{byun2026expert} because it trains its router with a second pre-trained model, a diffusion model fine-tuned on each domain.
Two controls of Lamarck isolate inheritance: the \emph{reset adapters} control re-initializes only the adapters for every domain, and the \emph{independent experts} control starts every expert from the first expert's initialization.
All Lamarck variants and controls use the readout of Eq.~\eqref{eq:readout}.

\paragraph{Protocol.}
For a fair comparison, all methods in Tab.~\ref{tab:system} are trained under one protocol: a frozen ViT-B/16 (ImageNet-21k$\rightarrow$1k), identical splits and domain orders, FP32, and five training seeds as in OS-Prompt~\cite{kim2024one} and CODA-Prompt~\cite{smith2023coda}.
Every method learns each domain for 15 epochs at batch size 128, the budget DUCT~\cite{zhou2025dual} releases for these benchmarks.
Lamarck and ten of the baselines use Adam with initial learning rate $10^{-3}$ for every trained parameter group, as released with CODA-Prompt, OS-Prompt, OS-Prompt++ and CPS-Prompt.
EASE and CONEC-LoRA instead use their released SGD settings.
Each method has its own schedule, weight decay and auxiliary phases (Tab.~\xref{tab:supp_baselines}).
On VAL, we select which tokens the adapters correct (Sec.~\xref{sec:supp_rank}), and we tune no learning rate on our splits for any method.
Lamarck uses the simpler of two readouts that differ by at most 0.2 points on VAL and TEST (Sec.~\xref{sec:supp_impl}).
We report TEST accuracy, with differences paired by seed and computed before rounding.

\subsection{Main results}
\label{sec:main}
\begin{figure}[t]
\centering
\includegraphics[width=\linewidth]{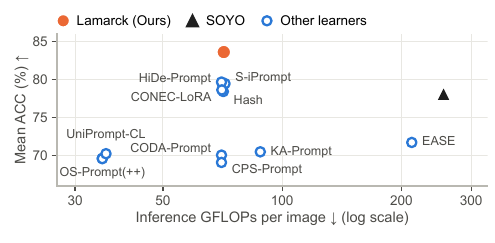}
\caption{\textbf{Lamarck has the highest mean ACC at the inference cost of a two-pass learner.} Mean ACC over the three benchmarks (Tab.~\ref{tab:system}) against GFLOPs per image on DN (log scale). Throughput in Tab.~\xref{tab:supp_eff} falls as GFLOPs rise (rank correlation $-0.94$). SOYO costs most: its released inference runs the ViT once per domain.}
\label{fig:efficiency}
\end{figure}
\input{sec/table_perdomain}
\paragraph{Lamarck is first on CDDB and in mean ACC.}
Lamarck has the best ACC and AIA on CDDB, 5.18 ACC points above Hash (Tab.~\ref{tab:system}).
Its mean ACC is also the highest: 83.62 against 79.67 for HiDe-Prompt, the best frozen-ViT baseline.
In ACC it is second on DN (69.46 against 69.91 for SOYO) and on OH (0.17 points below Hash).
The CDDB lead also holds with every TEST image weighted equally and with SAN removed (Tab.~\xref{tab:supp_cddb_views}).
With the CDDB order reversed, Lamarck reaches 93.15 ACC and leads Hash and HiDe-Prompt in every seed, on average by 4.70 and 6.54 points (Tab.~\xref{tab:supp_revorder}).
SLCA and DUCT, which fine-tune the ViT, lead on OH (90.14 and 88.87 vs.\ 88.47 ACC) but trail Lamarck on CDDB and DN (Tab.~\xref{tab:supp_duct}).
On DN, forgetting leaves the six shared-pool learners at least 8 ACC points below every per-domain learner: right after learning a domain, CODA-Prompt averages 71.3\%, as high as SOYO's 71.0\%, but ends at 50.6\% (Sec.~\xref{sec:supp_sharedpool}).

\paragraph{Lamarck costs as much as other two-pass learners.}
At test time, Lamarck needs 71.2 GFLOPs per image against 70.3--88.1 (Fig.~\ref{fig:efficiency}, Tab.~\xref{tab:supp_eff}).
Single-pass learners, all shared-pool, need 35--36 GFLOPs but trail Lamarck by over 13 points in mean ACC.
Lamarck stores 9.84M parameters (11.5\% of the backbone): each expert has its own prompt pool, adapters and readout.
EASE and CONEC-LoRA store more (Tab.~\xref{tab:supp_eff}).
\subsection{The adapters strengthen the experts}
\label{sec:gain}
\input{sec/table_modules}
\input{sec/table_variants}
\paragraph{Lamarck's lead comes from its experts.}
With all four banks routed identically, Lamarck's experts are the most accurate on OH and CDDB, under both routers and in DRA (Tab.~\ref{tab:router}).
Under the Mahalanobis router, their lead over SOYO's experts reaches 6.5 points (92.94 vs.\ 86.46 on CDDB).
Despite their adapters, SOYO's independently trained experts are less accurate than Lamarck's prompt-only experts on OH and CDDB.
On DN they are the most accurate (70.19 vs.\ 69.46 for Lamarck under the Mahalanobis router).

\paragraph{The adapters raise accuracy over Lamarck's prompt-only experts.}
Under the Mahalanobis router, the adapters add 4.36 ACC points on CDDB, 2.14 on DN and 0.33 on OH (Tabs.~\ref{tab:router} and~\ref{tab:modules}).
Of this gain, images routed to their own-domain expert contribute $+$4.75 points on CDDB, $+$2.00 on DN and $+$0.35 on OH, whereas misrouted images contribute between $-$0.39 and $+$0.14.
The adapters also raise accuracy on all three benchmarks under the prototype router and in DRA (Tab.~\ref{tab:router}).
They also lower SE on all three benchmarks, from 3.27\% to 1.80\% on CDDB (Tab.~\ref{tab:router}).
On DINOv2 ViT-B/14, they raise ACC in every seed on OH and CDDB (from 91.16 to 94.01 on CDDB; Tab.~\xref{tab:supp_dinov2}).

\paragraph{The adapters help where prompt-only experts fall short.}
In own-domain accuracy on DN, Lamarck's prompt-only experts are within 0.2 points of SOYO's adapter experts on real photographs but trail on the other five domains, from 1.5 points on painting to 8.2 on QuickDraw (Tab.~\ref{tab:perdomain}).
S-iPrompt's independently trained prompt-only experts share this shortfall: over the DN domains, they average 0.45 points below Lamarck's prompt-only experts (Tab.~\xref{tab:supp_perdomain}).
On DN, the adapters' gain follows exactly the order of this shortfall, from none on real photographs to 5.67 points on QuickDraw, where the gap to SOYO's experts narrows to 2.5 points.
The gain is 1.50 to 9.09 points across the CDDB domains but at most 0.93 on any OH domain.

\paragraph{Adapting every image token raises ACC over CLS-only adapters and deep-block prompts.}
Adapters that correct only the CLS token in the deep blocks give no ACC gain over Lamarck's prompt-only experts on OH and reach 92.30 on CDDB, against 92.94 for Lamarck (Tab.~\ref{tab:variants}).
On CDDB, adapting every image token also lowers the later-expert cost (15.75 vs.\ 23.53 for CLS-only adapters).
With as many parameters as the adapters, deep-block prompts recover 3.17 of the adapters' 4.36 points on CDDB.
The adapter form adds the other 1.19 points.
On OH it adds 0.25 (Tab.~\ref{tab:variants}).

\paragraph{Adapters in the deep blocks match the ACC of adapters in all twelve blocks; rank matters little.}
Adapters of rank 16 in all twelve blocks match Lamarck's ACC on DN, OH and CDDB (within 0.3 points; Tabs.~\ref{tab:variants} and~\xref{tab:supp_rank}).
Across ranks 8, 32 and 64, VAL and TEST ACC vary by at most 0.4 points on DN, 0.2 on OH and 1.1 on CDDB.
Selecting the VAL-best rank, $r{=}64$, would change TEST ACC by at most 0.4 points.

\paragraph{Prompt retrieval is nearly input-independent.}
In Lamarck's last prompt-only expert, the first prompted block's query (Eq.~\eqref{eq:retrieval}) is the learned class token plus a position embedding, identical for every image, so all images receive the same prompt.
In blocks 2--5, retrieval weights barely depend on the image.
On average, an image's weights differ from the dataset-average weights by less than 0.01 in total variation (at most 0.02 for any image), and their entropy is at least 98.4\% of the maximum on every benchmark (Tab.~\xref{tab:supp_retrieval}).
This expert therefore uses essentially one fixed prompt per block.
Querying every pool with the router's frozen feature makes retrieval input-dependent but changes OH ACC under the prototype router by at most 0.15 points (Tab.~\xref{tab:supp_sir}).
Moreover, when the deep blocks carry adapters, removing the prompts changes ACC by less than 0.25 points on OH and CDDB (Tab.~\ref{tab:modules}).
\subsection{Inheritance lowers the later-expert cost}
\label{sec:inherit}
\begin{figure}[t]
\centering
\includegraphics[width=0.97\linewidth]{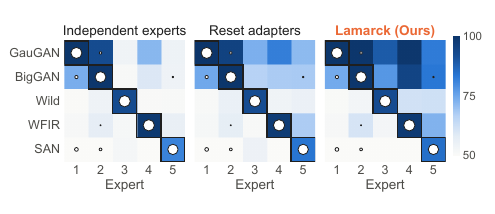}
\caption{\textbf{Inherited adapters make later experts more accurate on earlier domains.} Final-stage accuracy (colour, \%) of every expert (columns) on every CDDB domain (rows) in three expert banks. Above the diagonal: later experts on earlier domains. Outlined: own-domain experts. Dot area: share of a domain's images for which the Mahalanobis router selects each expert (same in all panels). Chance: 50\%; TEST, five-seed means.}
\label{fig:heatmap}
\end{figure}
\begin{figure}[t]
\centering
\includegraphics[width=\linewidth]{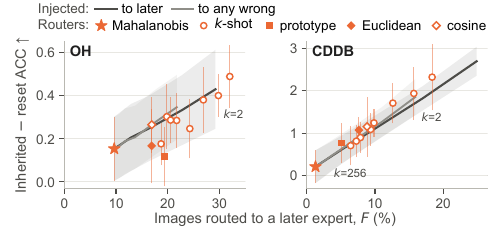}
\caption{\textbf{On OH and CDDB, inherited adapters add more ACC as more images reach later experts through router or injected errors.} TEST ACC: Lamarck minus reset adapters. Lines: up to 30\% of images moved from their Mahalanobis routes to a random later or any wrong expert. $k$-shot: the Mahalanobis router fitted on $k{=}2$ to 256 TRAIN images per domain. Euclidean, cosine: nearest TRAIN mean. Bands, bars: $\pm$1 s.d.\ over five seeds.}
\label{fig:inject}
\end{figure}
\input{sec/supp_costs}
\paragraph{Inheritance raises later-expert accuracy.}
On every benchmark, Lamarck has a lower later-expert cost than reset adapters and every bank without inheritance (Tabs.~\ref{tab:modules} and~\ref{tab:supp_costs}).
Without adapters, inheritance also lowers this cost, on CDDB from 35.49 to 13.58.
Adding reset adapters doubles it to 27.81, but inherited adapters raise it only to 15.75.
Lamarck's later-expert cost exceeds that of its prompt-only experts by 1.1--2.2 points on each benchmark, because adapters raise own-domain accuracy more than later-expert accuracy.
Over reset adapters, inheriting the adapters raises later-expert accuracy by 2.29, 1.47 and 12.06 points on DN, OH and CDDB (Tab.~\ref{tab:supp_costs}).
In Fig.~\ref{fig:heatmap}, every cell above the diagonal gains over reset adapters, whereas the other cells' mean changes by under half a point.
Inheritance still lowers the later-expert cost on reversed CDDB, from 34.25 to 27.85 (Tab.~\xref{tab:supp_revorder_inherit}), and on DINOv2 for OH and CDDB (Tab.~\xref{tab:supp_dinov2_inherit}).

\paragraph{Inheritance adds more ACC as more images reach later experts.}
Lamarck's ACC difference over reset adapters grows with $F$, the share of TEST images routed to a later expert, averaged over domains, for router and injected errors alike (Fig.~\ref{fig:inject}).
Under the Mahalanobis router, the difference is $+$0.15 on OH, $+$0.21 on CDDB and $-$0.41 on DN, where reset adapters have higher own-domain accuracy (Tab.~\xref{tab:supp_inherit_routing}).
On CDDB, fitting the router on $k$ TRAIN images per domain raises $F$ from 1.28\% to 18.37\% at $k{=}2$ and the difference to $+$2.32 points; it is positive in every seed for every $k$ (Tab.~\xref{tab:supp_kshot}).
Its slope across $k$, 0.129$\pm$0.050 points per point of $F$, is close to the 0.121 (12.06/100) implied by the 12.06-point gap in later-expert cost (Tab.~\ref{tab:modules}), and the OH and DN slopes are likewise close to their implied values (Tabs.~\ref{tab:supp_costs} and~\xref{tab:supp_kshot}).
When the last expert classifies every image, the difference is $+$1.32, $+$0.73 and $+$7.94 on DN, OH and CDDB (Tab.~\xref{tab:supp_inherit_routing}), whereas errors injected toward earlier experts do not raise it (Tab.~\xref{tab:supp_inject}).
Over independent experts, Lamarck's ACC lead is 0.08 to 0.58 points under the Mahalanobis router and 1.67 to 4.42 under the $k{=}2$ router (Tabs.~\xref{tab:supp_inherit_routing} and~\xref{tab:supp_kshot}).

%% file: sec/table_perdomain.tex
\begin{table}[t]
\centering
\caption{\textbf{The adapters raise own-domain accuracy most on QuickDraw and on CDDB.} Own-domain accuracy: $a_{j,j}$ in Eq.~\eqref{eq:diag}. Columns: SOYO's adapter experts, Lamarck's prompt-only experts and Lamarck. $\Delta$: gain from the adapters. TEST, five-seed mean$\pm$s.d. SAN has 33 TEST images.}
\label{tab:perdomain}
\scriptsize
\setlength{\tabcolsep}{6pt}
\renewcommand{\arraystretch}{1.1}
\begin{adjustbox}{max width=\linewidth}
\begin{tabular}{@{}llc@{\hskip 17pt}ccc@{}}
\toprule
 & & \multicolumn{4}{c}{Own-domain accuracy (\%) $\uparrow$} \\
\cmidrule(l){3-6}
 & Domain & SOYO & Prompts only & \textbf{Lamarck (Ours)} & $\Delta$ \\
\midrule
\multirow{6}{*}{\rotatebox[origin=c]{90}{DN}} & Clipart & $80.85{\scriptstyle\pm0.25}$ & $77.89{\scriptstyle\pm0.15}$ & $80.34{\scriptstyle\pm0.31}$ & $+2.46{\scriptstyle\pm0.29}$ \\
 & Infograph & $46.76{\scriptstyle\pm0.15}$ & $43.54{\scriptstyle\pm0.19}$ & $46.11{\scriptstyle\pm0.48}$ & $+2.58{\scriptstyle\pm0.55}$ \\
 & Painting & $76.08{\scriptstyle\pm0.18}$ & $74.54{\scriptstyle\pm0.23}$ & $75.55{\scriptstyle\pm0.12}$ & $+1.02{\scriptstyle\pm0.24}$ \\
 & QuickDraw & $72.29{\scriptstyle\pm0.18}$ & $64.12{\scriptstyle\pm0.11}$ & $69.79{\scriptstyle\pm0.09}$ & $+5.67{\scriptstyle\pm0.16}$ \\
 & Real & $87.41{\scriptstyle\pm0.04}$ & $87.21{\scriptstyle\pm0.19}$ & $87.17{\scriptstyle\pm0.09}$ & $-0.04{\scriptstyle\pm0.23}$ \\
 & Sketch & $73.47{\scriptstyle\pm0.14}$ & $71.00{\scriptstyle\pm0.16}$ & $72.64{\scriptstyle\pm0.17}$ & $+1.64{\scriptstyle\pm0.27}$ \\
\midrule
\multirow{4}{*}{\rotatebox[origin=c]{90}{OH}} & Art & $81.55{\scriptstyle\pm1.34}$ & $83.25{\scriptstyle\pm1.25}$ & $83.79{\scriptstyle\pm1.48}$ & $+0.53{\scriptstyle\pm0.50}$ \\
 & Clipart & $81.73{\scriptstyle\pm1.24}$ & $83.12{\scriptstyle\pm0.49}$ & $84.05{\scriptstyle\pm0.47}$ & $+0.93{\scriptstyle\pm0.53}$ \\
 & Product & $95.37{\scriptstyle\pm0.32}$ & $95.98{\scriptstyle\pm0.20}$ & $96.12{\scriptstyle\pm0.45}$ & $+0.14{\scriptstyle\pm0.40}$ \\
 & Real World & $91.40{\scriptstyle\pm0.46}$ & $93.31{\scriptstyle\pm0.46}$ & $93.19{\scriptstyle\pm0.42}$ & $-0.11{\scriptstyle\pm0.42}$ \\
\midrule
\multirow{5}{*}{\rotatebox[origin=c]{90}{CDDB}} & GauGAN & $99.78{\scriptstyle\pm0.10}$ & $98.08{\scriptstyle\pm0.48}$ & $99.58{\scriptstyle\pm0.11}$ & $+1.50{\scriptstyle\pm0.54}$ \\
 & BigGAN & $99.31{\scriptstyle\pm0.41}$ & $94.62{\scriptstyle\pm0.60}$ & $99.31{\scriptstyle\pm0.34}$ & $+4.69{\scriptstyle\pm0.58}$ \\
 & Wild & $87.84{\scriptstyle\pm0.64}$ & $87.16{\scriptstyle\pm1.05}$ & $92.29{\scriptstyle\pm0.34}$ & $+5.13{\scriptstyle\pm1.19}$ \\
 & WFIR & $72.00{\scriptstyle\pm5.85}$ & $91.12{\scriptstyle\pm1.20}$ & $97.88{\scriptstyle\pm0.71}$ & $+6.75{\scriptstyle\pm1.35}$ \\
 & SAN & $81.82{\scriptstyle\pm0.00}$ & $75.76{\scriptstyle\pm4.79}$ & $84.85{\scriptstyle\pm2.14}$ & $+9.09{\scriptstyle\pm6.78}$ \\
\bottomrule
\end{tabular}
\end{adjustbox}
\end{table}

%% file: sec/table_modules.tex
\begin{table}[t]
\centering
\caption{\textbf{Removing the adapters lowers ACC most on CDDB, and removing inheritance raises the later-expert cost.} Top row: Lamarck. Other rows: per-seed change after removing components. ACC: Mahalanobis router. Later-expert cost: Eq.~\eqref{eq:costs}. TEST, five-seed mean$\pm$s.d.}
\label{tab:modules}
\scriptsize
\setlength{\tabcolsep}{2.2pt}
\begin{adjustbox}{max width=\linewidth}
\begin{tabular}{@{}lcccc@{}}
\toprule
 & \multicolumn{2}{c}{ACC$\uparrow$} & \multicolumn{2}{c}{Later-expert cost$\downarrow$} \\
\cmidrule(lr){2-3}\cmidrule(l){4-5}
Expert bank & OH & CDDB & OH & CDDB \\
\midrule
\textbf{Lamarck (Ours)} & $88.47{\scriptstyle\pm0.25}$ & $92.94{\scriptstyle\pm0.62}$ & $5.04{\scriptstyle\pm0.48}$ & $15.75{\scriptstyle\pm0.84}$ \\
\midrule
w/o adapters & $-0.33{\scriptstyle\pm0.19}$ & $-4.36{\scriptstyle\pm1.27}$ & $-1.08{\scriptstyle\pm0.45}$ & $-2.18{\scriptstyle\pm1.76}$ \\
w/o prompts & $+0.16{\scriptstyle\pm0.28}$ & $-0.23{\scriptstyle\pm0.70}$ & $+0.20{\scriptstyle\pm0.48}$ & $-0.48{\scriptstyle\pm2.43}$ \\
w/o adapter inheritance (reset) & $-0.15{\scriptstyle\pm0.15}$ & $-0.21{\scriptstyle\pm0.39}$ & $+1.51{\scriptstyle\pm0.63}$ & $+12.06{\scriptstyle\pm1.62}$ \\
w/o inheritance (independent) & $-0.54{\scriptstyle\pm0.28}$ & $-0.58{\scriptstyle\pm0.75}$ & $+8.37{\scriptstyle\pm0.78}$ & $+22.78{\scriptstyle\pm1.43}$ \\
w/o adapters and inheritance & $-0.95{\scriptstyle\pm0.19}$ & $-3.51{\scriptstyle\pm0.66}$ & $+7.55{\scriptstyle\pm0.54}$ & $+19.74{\scriptstyle\pm0.94}$ \\
\bottomrule
\end{tabular}
\end{adjustbox}
\end{table}

%% file: sec/table_variants.tex
\begin{table}[t]
\centering
\caption{\textbf{Adapters on every image token give higher ACC than CLS-only adapters or prompts in the deep blocks.} Each variant changes only Lamarck's adapters ($r$: rank) and still uses inheritance. The last row replaces the adapters with 64 prompt tokens in each deep block, matching their parameter count. Columns as in Tab.~\ref{tab:modules}. TEST, five-seed mean$\pm$s.d.}
\label{tab:variants}
\scriptsize
\setlength{\tabcolsep}{2.2pt}
\begin{adjustbox}{max width=\linewidth}
\begin{tabular}{@{}lcccc@{}}
\toprule
 & \multicolumn{2}{c}{ACC$\uparrow$} & \multicolumn{2}{c}{Later-expert cost$\downarrow$} \\
\cmidrule(lr){2-3}\cmidrule(l){4-5}
Variant & OH & CDDB & OH & CDDB \\
\midrule
\textbf{Lamarck (Ours)}, $r{=}32$ & $88.47{\scriptstyle\pm0.25}$ & $92.94{\scriptstyle\pm0.62}$ & $5.04{\scriptstyle\pm0.48}$ & $15.75{\scriptstyle\pm0.84}$ \\
CLS token only ($r{=}64$) & $88.15{\scriptstyle\pm0.63}$ & $92.30{\scriptstyle\pm0.72}$ & $4.60{\scriptstyle\pm0.89}$ & $23.53{\scriptstyle\pm1.52}$ \\
Blocks 1--12 ($r{=}16$) & $88.54{\scriptstyle\pm0.40}$ & $93.20{\scriptstyle\pm0.97}$ & $5.41{\scriptstyle\pm0.66}$ & $13.17{\scriptstyle\pm0.78}$ \\
Deep prompts, no adapters & $88.22{\scriptstyle\pm0.26}$ & $91.74{\scriptstyle\pm0.19}$ & $4.96{\scriptstyle\pm0.28}$ & $16.59{\scriptstyle\pm2.63}$ \\
\bottomrule
\end{tabular}
\end{adjustbox}
\end{table}

%% file: sec/supp_costs.tex
\begin{table*}[t]
\centering
\caption{\textbf{The two fully inherited banks have the highest later-expert accuracy and the lowest later-expert cost.} Metrics as defined in Eq.~\eqref{eq:costs}. Experts: what each expert inherits from its predecessor. TEST, five-seed mean$\pm$s.d.}
\label{tab:supp_costs}
\scriptsize
\setlength{\tabcolsep}{2.4pt}
\begin{adjustbox}{max width=\textwidth}
\begin{tabular}{@{}llccccccccc@{}}
\toprule
 &  & \multicolumn{3}{c}{Later-expert accuracy$\uparrow$} & \multicolumn{3}{c}{Later-expert cost$\downarrow$} & \multicolumn{3}{c}{Earlier-expert cost$\downarrow$} \\
\cmidrule(lr){3-5}\cmidrule(lr){6-8}\cmidrule(l){9-11}
Expert bank & Experts & DN & OH & CDDB & DN & OH & CDDB & DN & OH & CDDB \\
\midrule
S-iPrompt & independent & $31.52\,{\scriptstyle\pm\,0.11}$ & $73.22\,{\scriptstyle\pm\,0.21}$ & $57.64\,{\scriptstyle\pm\,1.05}$ & $35.19\,{\scriptstyle\pm\,0.19}$ & $12.22\,{\scriptstyle\pm\,0.45}$ & $33.38\,{\scriptstyle\pm\,1.42}$ & $38.77\,{\scriptstyle\pm\,0.12}$ & $12.31\,{\scriptstyle\pm\,0.51}$ & $30.11\,{\scriptstyle\pm\,5.15}$ \\
SOYO & independent & $33.07\,{\scriptstyle\pm\,0.13}$ & $66.78\,{\scriptstyle\pm\,0.58}$ & $54.47\,{\scriptstyle\pm\,0.55}$ & $37.04\,{\scriptstyle\pm\,0.15}$ & $17.13\,{\scriptstyle\pm\,1.25}$ & $40.00\,{\scriptstyle\pm\,0.97}$ & $40.02\,{\scriptstyle\pm\,0.19}$ & $17.27\,{\scriptstyle\pm\,0.58}$ & $30.24\,{\scriptstyle\pm\,1.95}$ \\
Prompts only, independent & independent & $30.28\,{\scriptstyle\pm\,0.22}$ & $72.48\,{\scriptstyle\pm\,0.40}$ & $58.61\,{\scriptstyle\pm\,0.87}$ & $36.48\,{\scriptstyle\pm\,0.18}$ & $12.59\,{\scriptstyle\pm\,0.30}$ & $35.49\,{\scriptstyle\pm\,0.70}$ & $40.53\,{\scriptstyle\pm\,0.26}$ & $13.40\,{\scriptstyle\pm\,0.91}$ & $34.89\,{\scriptstyle\pm\,1.88}$ \\
Lamarck, independent & independent & $30.85\,{\scriptstyle\pm\,0.16}$ & $72.31\,{\scriptstyle\pm\,0.59}$ & $59.45\,{\scriptstyle\pm\,1.00}$ & $38.52\,{\scriptstyle\pm\,0.19}$ & $13.42\,{\scriptstyle\pm\,0.84}$ & $38.53\,{\scriptstyle\pm\,1.10}$ & $41.20\,{\scriptstyle\pm\,0.15}$ & $13.71\,{\scriptstyle\pm\,0.89}$ & $37.81\,{\scriptstyle\pm\,1.48}$ \\
Lamarck, reset adapters & prompts inherited & $35.54\,{\scriptstyle\pm\,0.33}$ & $79.42\,{\scriptstyle\pm\,0.48}$ & $70.06\,{\scriptstyle\pm\,1.41}$ & $34.02\,{\scriptstyle\pm\,0.30}$ & $6.55\,{\scriptstyle\pm\,0.82}$ & $27.81\,{\scriptstyle\pm\,1.42}$ & $39.34\,{\scriptstyle\pm\,0.14}$ & $12.13\,{\scriptstyle\pm\,1.08}$ & $37.84\,{\scriptstyle\pm\,1.49}$ \\
Lamarck, prompts only & inherited & $37.28\,{\scriptstyle\pm\,0.13}$ & $81.37\,{\scriptstyle\pm\,0.25}$ & $80.59\,{\scriptstyle\pm\,1.19}$ & $29.56\,{\scriptstyle\pm\,0.08}$ & $3.96\,{\scriptstyle\pm\,0.73}$ & $13.58\,{\scriptstyle\pm\,1.24}$ & $37.70\,{\scriptstyle\pm\,0.10}$ & $11.85\,{\scriptstyle\pm\,0.84}$ & $32.33\,{\scriptstyle\pm\,2.30}$ \\
\textbf{Lamarck (Ours)} & inherited & $37.83\,{\scriptstyle\pm\,0.13}$ & $80.89\,{\scriptstyle\pm\,0.19}$ & $82.12\,{\scriptstyle\pm\,0.87}$ & $31.47\,{\scriptstyle\pm\,0.18}$ & $5.04\,{\scriptstyle\pm\,0.48}$ & $15.75\,{\scriptstyle\pm\,0.84}$ & $37.65\,{\scriptstyle\pm\,0.16}$ & $11.93\,{\scriptstyle\pm\,0.74}$ & $38.93\,{\scriptstyle\pm\,1.46}$ \\
\bottomrule
\end{tabular}
\end{adjustbox}
\end{table*}

%% file: sec/6_conclusion.tex
\section{Conclusion}
\label{sec:conclusion}
Per-domain learners select one expert for each test image with a router, and recent work trains this router.
For the learners we analyze, a training-free Mahalanobis router costs under 2.7 ACC points; on each benchmark, this is less than the largest ACC difference between their expert banks, so the experts limit accuracy.
Lamarck uses this router and strengthens the experts.
Its adapters correct every image token after each deep block.
Nine of our twelve baselines train nothing in these blocks.
Each expert is initialized from its predecessor, which adds no parameters or computation and makes later experts more accurate on earlier domains.
Inheritance thus adds more ACC as more images reach later experts.
Against twelve frozen-backbone baselines, Lamarck is first on CDDB by 5.2 points, second on DN and OH, and highest in mean ACC.

\paragraph{Limitations.}
On DN, SOYO's experts are more accurate than Lamarck's on every domain, and SOYO's ACC is higher.
Under the Mahalanobis router, inheritance changes ACC against reset adapters by $-$0.41 to $+$0.21 points; its gain becomes clear only under weaker routers or injected routing errors.
Without adapters, inherited experts trail independent ones in ACC on CDDB (Tab.~\ref{tab:modules}).
The second backbone, DINOv2 ViT-B/14, is tested only on OH and CDDB, and the domain order is reversed only on CDDB.
Storage exceeds that of the other prompt learners and grows by one expert per domain; routing adds an unprompted pass.

%% file: sec/llm_usage.tex
\section*{LLM Usage Statement}
\label{sec:llm_usage}

The authors used OpenAI tools (ChatGPT and Codex) and Anthropic tools (Claude and Claude Code) to support research and manuscript preparation.
The research hypotheses were formulated by the authors based on limitations they identified in prior work.
The tools supported the critical examination of these hypotheses and the proposed methodology, providing suggestions and feedback on the methodological rationale and the empirical support for the reported claims.
They also facilitated experimental implementation through code generation and refactoring, including routines for data loading, preprocessing, and dataset splitting.
During manuscript preparation, the tools assisted with English translation, language revision, preparation of visual material, and the organization and refinement of research and related work material prepared by the authors.

The authors critically assessed the outputs and suggestions provided by these tools and incorporated or revised them as appropriate.
All final scientific decisions, including those concerning the research direction, methodology, experimental evaluation, and interpretation of results, were made by the authors.
The authors reviewed the technical implementation and complete manuscript and take full responsibility for the accuracy and integrity of the submitted work.

%% file: sec/supp.tex
\raggedbottom%
\makeatletter\setlength{\@fptop}{0pt}\setlength{\@dblfptop}{0pt}\setlength{\@fpsep}{16pt}\setlength{\@dblfpsep}{16pt}\makeatother%
\makeatletter
\DeclareRobustCommand{\suppsecdot}{\hskip -1em.~}
\def\cvprsect#1{\cvprsection{\texorpdfstring{\suppsecdot}{}#1}\addcontentsline{stc}{suppsec}{\protect\numberline{\thesection}#1}}
\newcommand*\l@suppsec{\@dottedtocline{1}{0em}{2.1em}}
\newcommand{\printsupptoc}{\begingroup\small\setlength{\parskip}{0pt}\@starttoc{stc}\endgroup}
\makeatother
\setcounter{section}{0}
\renewcommand{\thesection}{S\arabic{section}}
\renewcommand{\theHsection}{S\arabic{section}}%
\setcounter{table}{0}
\renewcommand{\thetable}{S\arabic{table}}
\setcounter{figure}{0}
\renewcommand{\thefigure}{S\arabic{figure}}
\setcounter{algorithm}{0}
\renewcommand{\thealgorithm}{S\arabic{algorithm}}
\maketitlesupplementary
\printsupptoc

As in the main paper, ACC and AIA are the final and the average incremental accuracy (Eq.~\xeqref{eq:metrics}), and TRAIN, VAL and TEST are the training, validation and test splits.
Tables, figures, sections and equations without the prefix S are those of the main paper.

\section{Implementation details}
\label{sec:supp_impl}
All learners use the same ViT-B/16 timm checkpoint:\newline\texttt{vit\_base\_patch16\_224.augreg\_in21k\_ft\_in1k}.
\paragraph{Lamarck.}
Lamarck follows the protocol of Sec.~\xref{sec:setup} with a cosine learning-rate schedule, and the last epoch is kept.
In DomainNet, where $T{=}6$ does not divide $Z$, the last stage also receives the four remaining rows.
In independent experts, the prompt rows of earlier stages stay visible at their initial values.
Two readouts were compared on prompt-only experts: the adopted one (Eq.~\xeqref{eq:readout}) and one with an additional cosine classifier and cosine loss.
They differ by at most 0.2 points in ACC and AIA on VAL and TEST.
Algorithm~\ref{alg:lamarck} summarizes training and inference.

\begin{algorithm}[t]
\caption{Lamarck: training and inference}
\label{alg:lamarck}
\small
\begin{algorithmic}[1]
\Require domains $\mathcal{D}_1,\dots,\mathcal{D}_T$ in sequence; frozen backbone $\Theta$
\Statex \textit{Training}
\For{$t=1,\dots,T$}
  \State \textbf{if} $t{>}1$: $\psi_t\gets\psi_{t-1}$ \textbf{else} $U_b\gets0$ \Comment{inheritance}
  \State add the prompt rows of stage $t$ to the pool
  \State train them, all $\{D_b,U_b\}$ and $(f_{\mathrm{fc}},w)$ on $\mathcal{D}_t$ \Comment{$\Theta$ frozen}
  \State freeze and store $\psi_t$
  \State $\mu_t\gets$ mean of $z(x)$ over $\mathcal{D}_t$; update $S$ and $\Sigma$
\EndFor
\Statex \textit{Inference on a test image $x$ after stage $t$}
\State $z\gets z(x)$ \Comment{unprompted pass}
\State $e^{\star}\gets\arg\min_{e\le t}\,(z-\mu_e)^{\top}\Sigma^{-1}(z-\mu_e)$ \Comment{Eq.~\xeqref{eq:router}}
\State \Return $\arg\max_y \mathrm{logits}(x)_y$ of $\psi_{e^{\star}}$ \Comment{prompted pass}
\end{algorithmic}
\end{algorithm}

\paragraph{Splits.}
DomainNet uses the official test set as TEST and holds out 10\% of the official training set as VAL.
OfficeHome, which has no official split, is divided at random into 70/14/16\% TRAIN/VAL/TEST.
For CDDB-Hard we pool the official training and validation images of each domain and re-split them 80/10/10\%, keeping duplicate images within one split.
All methods and seeds share these splits and the same preprocessing.

\paragraph{Hardware.}
Lamarck and its prompt-only experts were trained on a V100 (PyTorch 2.4.1) and an RTX PRO 6000 (PyTorch 2.9.1).
The GPU has little effect: it changes the adapter gain by $-$0.02 points on OfficeHome (Mahalanobis router) and $-$0.27 on CDDB-Hard (prototype router), and Lamarck's OfficeHome ACC by 0.08 points (88.39$\pm$0.20 on the RTX PRO 6000, 88.47$\pm$0.25 on the V100).

\section{Baseline settings}
\label{sec:supp_settings}
For a fair comparison, every baseline of Tab.~\xref{tab:system} runs inside one harness that fixes the backbone, the splits, the domain order and the per-domain budget; Tab.~\ref{tab:supp_baselines} lists the settings of each method.
The harness also fixes the prompt budget of the prompt learners at the values released with CODA-Prompt, OS-Prompt and CPS-Prompt: prompts of 8 tokens (4 key and 4 value tokens for prefix prompts) and, where a method keeps a pool, 100 prompts; UniPrompt-CL keeps its released pool.
The Adam setting of Sec.~\xref{sec:setup} also trains SOYO's router, and SLCA and DUCT, which fine-tune the backbone, keep their released SGD settings within the same budget.
SOYO runs the model, loss and inference code of its release.
The all-expert evaluation of Sec.~\xref{sec:diagnosis} reproduces each learner's own routed predictions exactly in every evaluated cell.

\input{sec/supp_baselines}

\section{Own-domain accuracy of S-iPrompt}
S-iPrompt is within 0.35 points of Lamarck's prompt-only experts on every DomainNet domain except Sketch (1.78 points; Tab.~\ref{tab:supp_perdomain}).
\input{sec/supp_perdomain_all}

\section{Prompt retrieval}
Table~\ref{tab:supp_retrieval} gives, per prompted block, the retrieval statistics summarized in Sec.~\xref{sec:gain}.
For Tab.~\ref{tab:supp_sir}, every pool is queried with the router's frozen feature: $Q_i$ in Eq.~\xeqref{eq:retrieval} is replaced by $z(x)-\bar\mu$, with $\bar\mu$ the mean of the first domain's router prototypes, and $\gamma$ by $\gamma/\tau$, with $\tau{=}0.3$ chosen on VAL; the sharpest retrieval ($\tau{=}0.03$) has the lowest VAL ACC.
\input{sec/supp_retrieval}
\input{sec/supp_sir}

\section{Fine-tuned backbones: SLCA and DUCT}
SLCA~\cite{zhang2023slca} fine-tunes the backbone with a reduced learning rate and aligns the classifier with class-wise feature statistics after each stage.
Averaged over the three benchmarks, SLCA and DUCT reach 82.49 and 72.14 ACC, against 83.62 for Lamarck (Tab.~\ref{tab:supp_duct}).
\input{sec/supp_table_duct}

\section{Other views of CDDB-Hard}
Table~\ref{tab:supp_cddb_views} adds to the CDDB-Hard ACC of Tab.~\xref{tab:system} three other aggregates and the accuracy on each domain; Lamarck is also first on every domain except SAN.
\input{sec/supp_cddb_views}
With the domains learned in reverse (SAN to GauGAN), Lamarck reaches 93.15 ACC, against 92.94 in the original order, and leads Hash and HiDe-Prompt in every seed, by 4.70 and 6.54 points (Tab.~\ref{tab:supp_revorder}); its AIA in that order is also above theirs (90.12 vs.\ 86.30 and 83.71).
\input{sec/supp_revorder}

\section{Baselines on DomainNet}
\label{sec:supp_sharedpool}
On DomainNet, the gap between shared-pool and per-domain learners is forgetting, not weaker training.
Averaged over the six domains and five seeds, the accuracy on a domain right after learning it is 71.34\% for CODA-Prompt and 68.36--71.20\% for OS-Prompt, OS-Prompt++, KA-Prompt, CPS-Prompt and UniPrompt-CL (UniPrompt-CL over four seeds), against 71.01\% for SOYO, 69.46\% for CONEC-LoRA, 68.87\% for Hash, 67.03\% for S-iPrompt, 66.19\% for HiDe-Prompt and 64.15\% for EASE.
After the last domain, the six shared-pool learners keep at most 50.62\% (Tab.~\xref{tab:system}); CODA-Prompt's largest losses are on QuickDraw (68.39 to 21.22) and Infograph (45.02 to 24.40).
Lamarck is more accurate than EASE on every domain and than CONEC-LoRA on four of six, most on QuickDraw (69.76\% vs.\ 50.32\% and 60.66\%), while CONEC-LoRA leads on Infograph (44.92\% vs.\ 44.16\%) and Painting (73.35\% vs.\ 70.79\%; Tab.~\ref{tab:supp_dn_perdomain}).
\input{sec/supp_dn_perdomain}

\section{Test-time and training cost}
\label{sec:supp_cost}
Table~\ref{tab:supp_eff} gives the cost behind Fig.~\xref{fig:efficiency}, measured on one Tesla V100 (32\,GB) in FP32 with batch size 64 and synthetic $224{\times}224$ inputs, each method in its released environment.
EASE passes each test image through its six adapter stacks; SOYO's released inference runs the encoder once per domain and keeps the output of the selected expert.
GFLOPs per image are counted with PyTorch's FLOP counter (matrix products, convolutions and attention; two FLOPs per multiply--add), and training GFLOPs include the backward pass.
Param.\ excludes the 85.80M parameters of the backbone.
Inference throughput is the mean over timed repeats (s.d.\ at most 0.90 images/s); Lamarck splits each batch evenly over its six experts and reaches 160.3 images/s when one expert serves the whole batch.
A training step is one optimizer step of the last stage, timed over three repeats (s.d.\ at most 0.35 images/s) on a V100 of the same machine; Lamarck trains with one pass, through the expert being trained.
Memory is the peak allocated by PyTorch.
\input{sec/supp_eff}

\section{Adapter rank and placement}
\label{sec:supp_rank}
Lamarck's accuracy depends little on the rank of its adapters (Tab.~\ref{tab:supp_rank}): among $r{=}8$, 32 (Lamarck's rank) and 64, the rank that is best on VAL, $r{=}64$ on all three benchmarks, changes TEST ACC by at most 0.4 points.
VAL selects the adapted tokens: on DomainNet's first domain, clipart (one seed), the first expert reaches an own-domain VAL accuracy of 80.04 with adapters on every image token after blocks 6--12 (Lamarck) and 78.43 when they correct only the CLS token ($r{=}64$), against 77.54 as a prompt-only expert.
Adapters in all twelve blocks ($r{=}16$) reach 80.61 in the same setting, 0.57 points higher; we keep blocks 6--12 so that prompts and adapters occupy separate blocks, and report the twelve-block adapters as a control.
With Lamarck's prompts and inheritance kept, they are within 0.3 TEST ACC points of Lamarck on all three benchmarks; on DomainNet they reach 69.49, below SOYO's 69.91 with its own router and 70.19 under the Mahalanobis router (Tabs.~\xref{tab:system} and~\xref{tab:router}).
Against deep-block prompts with as many parameters (Tab.~\xref{tab:variants}), the adapter form leads in all five seeds on CDDB-Hard ($+$1.19 points) and in four of five on OfficeHome ($+$0.25).
On CDDB-Hard, the gap between VAL and TEST ACC comes almost entirely from SAN (33 images per split); on the other four domains, VAL and TEST accuracy agree within 1.6 points.
\input{sec/supp_rank}

\section{Further results on inheritance}
\begin{figure}[t]
\centering
\includegraphics[width=\linewidth]{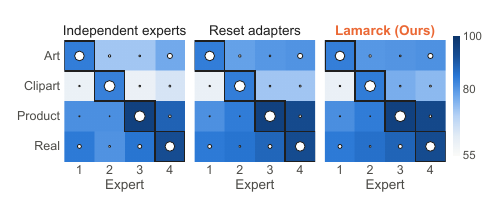}
\caption{\textbf{Figure~\xref{fig:heatmap} on OfficeHome.} Panels, cells and dot areas as in Fig.~\xref{fig:heatmap}; TEST, five-seed means.}
\label{fig:supp_heatmap_oh}
\end{figure}

\paragraph{OfficeHome.}
In Fig.~\ref{fig:supp_heatmap_oh}, Lamarck is 8.58 points more accurate than independent experts above the diagonal (every cell and seed) and 2.04 points more accurate below it, so on OfficeHome inheritance helps mainly, but not only, the later experts.

\begin{figure}[t]
\centering
\includegraphics[width=\linewidth]{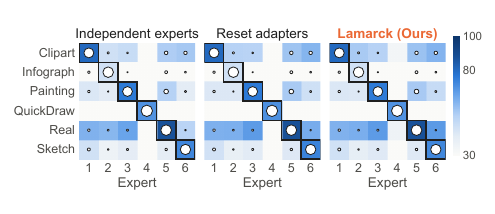}
\caption{\textbf{Figure~\xref{fig:heatmap} on DomainNet.} Panels, cells and dot areas as in Fig.~\xref{fig:heatmap}, with the colour scale starting at 30\%; TEST, five-seed means.}
\label{fig:supp_heatmap_dn}
\end{figure}

\paragraph{DomainNet.}
In Fig.~\ref{fig:supp_heatmap_dn}, Lamarck is 6.98 points more accurate than independent experts above the diagonal (every cell and seed) and 3.47 points more accurate below it.
Against reset adapters, it is 2.29 points more accurate above the diagonal and 1.13 points below it, so on DomainNet, too, inheritance helps mainly the later experts.
The router misroutes 11.9\% of DomainNet's TEST images but only 0.1\% of QuickDraw images, on which every other expert reaches at most 17.0\% accuracy.
Of the misrouted images, 63\% either come from Real, the largest domain, or are sent to its expert; the most frequent case is a Painting image sent to the Real expert (17.5\% of all misroutes).
On the misrouted images, Lamarck's selected experts reach 60.1\% accuracy, against 76.9\% for the own-domain experts.

\paragraph{Routing conditions.}
Table~\ref{tab:supp_inherit_routing} completes the differences of Sec.~\xref{sec:inherit} with domain-label routing, the share of misrouted images and the s.d.\ over seeds.
On DomainNet, inheritance lowers both the later- and the earlier-expert cost against reset adapters in every seed (Tab.~\xref{tab:supp_costs}), yet its ACC under the Mahalanobis router is 69.46 against 69.87, because reset adapters are 0.41 points more accurate on their own domains, in all five seeds, and the router loses 2.47 points with either bank (Tabs.~\xref{tab:router} and~\ref{tab:supp_inherit_routing}).
On CDDB-Hard, inheritance lowers the later-expert cost but not the earlier-expert cost (38.93 vs.\ 37.84 for reset adapters; Tab.~\xref{tab:supp_costs}).
\input{sec/supp_inherit_routing}

\begin{figure}[t]
\centering
\includegraphics[width=\linewidth]{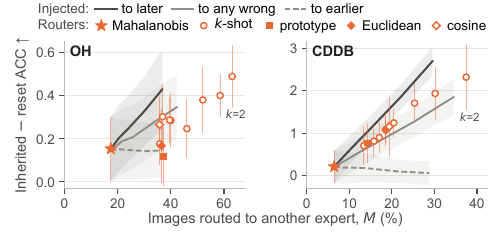}
\caption{\textbf{Against the share of all misrouted images, routers and injected errors align less well than against $F$.} Fig.~\xref{fig:inject} with $M$, the share of TEST images routed to any other expert (mean over domains), on the horizontal axis, and with errors injected toward earlier experts (dashed).}
\label{fig:supp_inject_M}
\end{figure}

\paragraph{Routing-error injection.}
Table~\ref{tab:supp_inject} gives the values behind the lines of Fig.~\xref{fig:inject}, together with errors injected toward earlier experts, and repeats the injection, with the same draws, against independent experts and for prompt-only experts.
Against independent experts too, the advantage of inheritance grows with $p$ when the extra errors reach later experts and grows far less, or not at all, when they reach earlier experts.
Inherited prompt-only experts trail independent ones on CDDB-Hard at the router's own error rate, $p{=}0$ ($-$0.85 points), but lead at $p{=}30\%$ by 4.09 points when the extra errors go to later experts and by 2.76 when they go to any wrong expert.
When the last expert classifies every image, Lamarck reaches 48.38\% on DomainNet, against 47.06\% with reset adapters and 42.21\% with independent experts; the bottom rows of Tab.~\ref{tab:supp_inject} give these values on OfficeHome and CDDB-Hard.
Fig.~\xref{fig:inject} draws each injected share $p$ at the $F$ it gives in expectation from each seed's Mahalanobis routes; errors injected toward earlier experts do not raise $F$ and are drawn against $M$ in Fig.~\ref{fig:supp_inject_M}.
\input{sec/supp_inject}
\input{sec/supp_kshot}
\input{sec/supp_revorder_inherit}
\input{sec/supp_revorder_pairs}

\paragraph{Routers that make their own errors.}
Routers that send more TEST images to later experts raise the advantage of inheritance as injected errors do (Tab.~\ref{tab:supp_kshot}, markers of Fig.~\xref{fig:inject}).
All routers are training-free, use only TRAIN statistics of the frozen feature $z(x)$ of Eq.~\xeqref{eq:router} and give every expert bank the same routes: the Mahalanobis router fitted on all TRAIN images or on $k\in\{2,4,8,16,32,64,128,256\}$ TRAIN images per domain, drawn without replacement (20 draws per seed and $k$, averaged within each seed), Euclidean and cosine nearest-mean routers, the prototype router of Sec.~\xref{sec:diagnosis} and the last expert alone.
The routers, the $k$ grid and the slope comparison were fixed before any expert bank was scored on TEST.
On CDDB-Hard, every router that sends more images to later experts than the Mahalanobis router gives Lamarck a lead over reset adapters in all five seeds; under the $k$-shot routers it rises from $+$0.70 at $k{=}256$ to $+$2.32 at $k{=}2$.
Images routed to their own expert contribute $+$0.24 to $+$0.42 under every router, so the growth comes from images at later experts ($+$0.05 under the Mahalanobis router, $+$2.06 at $k{=}2$).
The least-squares slope of the lead on $F$, fitted per seed over the $k$-shot draws and the Mahalanobis router, is 0.129$\pm$0.050 points per point of $F$ (0.072 to 0.201 across seeds), against 0.105 for errors injected toward later experts and 0.121 predicted by the 12.06-point gap in later-expert cost (Tab.~\xref{tab:modules}); three of five seeds lie at or above both.
On OfficeHome, the lead is $+$0.18 at $k{=}256$ and $+$0.49 at $k{=}2$, positive in all five seeds at every $k$, and its slope, 0.016$\pm$0.006, is close to the 0.015 that the 1.51-point gap in later-expert cost predicts.
On DomainNet, the difference from reset adapters also rises with $F$, by 0.031$\pm$0.006 points per point of $F$ (0.022 to 0.037 across seeds), close to the 0.026 implied by the 2.55-point gap in later-expert cost (Tab.~\xref{tab:supp_costs}).
It is $-$0.41 under the Mahalanobis router ($F{=}9.18\%$) and $-$0.13 to $-$0.36 under every other router except the $k{=}2$ router ($+$0.13, $F{=}20.70\%$, four of five seeds) and the last expert ($+$1.32); images routed to their own expert lower it by 0.15 to 0.38 points under every router, as reset adapters are more accurate on their own domains.
Against independent experts, the lead grows from $+$1.18 to $+$4.42 on CDDB-Hard, from $+$1.27 to $+$3.07 on OfficeHome and from $+$0.44 to $+$1.67 on DomainNet between $k{=}256$ and $k{=}2$ (slopes 0.259$\pm$0.049, 0.108$\pm$0.018 and 0.105$\pm$0.004); on DomainNet it is $+$0.08 under the Mahalanobis router (four of five seeds) and positive in every seed under every other router.
On CDDB-Hard, the lead over reset adapters rises alike when ACC is taken over all TEST images instead of the domain mean or without SAN, from $+$0.09 and $+$0.11 under the Mahalanobis router to $+$2.67 and $+$2.82 at $k{=}2$, and is positive under every router.
At $k{=}128$, where the pooled covariance is fitted from 640 (CDDB-Hard) and 512 (OfficeHome) TRAIN images in 768 dimensions, $F$ exceeds that at $k{=}64$ (9.36\% vs.\ 7.27\% and 20.56\% vs.\ 19.82\%), and so does the CDDB-Hard lead ($+$1.09 vs.\ $+$0.81).
Against $M$, the share of images routed to any other expert, routers and injected errors align less well (Fig.~\ref{fig:supp_inject_M}), because the routers also send many images to earlier experts (31.1\% on OfficeHome at $k{=}2$), which change the lead little ($+$0.09).

\paragraph{Reversed domain order.}
With the CDDB-Hard domains learned from SAN to GauGAN, inheritance still lowers the later-expert cost in every seed, by 6.40$\pm$1.15 points (27.85 vs.\ 34.25) instead of 12.06$\pm$1.62, and adds 3.29 instead of 7.94 points when the last expert classifies every image, also in every seed (Tab.~\ref{tab:supp_revorder_inherit}).
Under the Mahalanobis router it changes ACC by $+$0.03$\pm$0.26 (two of five seeds positive) instead of $+$0.21$\pm$0.39, under the prototype router by $+$0.34 instead of $+$0.77, DRA by $-$0.10 instead of $+$0.31, and the earlier-expert cost by $+$0.83 instead of $+$1.08.
On the same domain pairs, an expert trained after the image's domain costs less than one trained before it, and inheritance enlarges this drop 2.5 to 3.1 times (Tab.~\ref{tab:supp_revorder_pairs}).
The two orders give each pair (image domain, expert domain) once with the expert trained after the image's domain and once before it; with inherited adapters, the drop is 21.74 points on the ten pairs whose expert follows the image's domain in the original order (U) and 11.08 on the ten converse pairs (L), against 8.85 and 3.60 with reset adapters, positive in every seed.

\section{Second backbone: DINOv2 ViT-B/14}
\label{sec:supp_dinov2}
On DINOv2 ViT-B/14, Lamarck runs on OfficeHome and CDDB-Hard with the configuration used on ViT-B/16, without any tuning on DINOv2, over five seeds.
Its adapters raise ACC over Lamarck's prompt-only experts in every seed, from 88.56 to 89.34 on OfficeHome and from 91.16 to 94.01 on CDDB-Hard (Tab.~\ref{tab:supp_dinov2}).
Against reset adapters, inheritance lowers the later-expert cost (5.93 vs.\ 9.01 on OfficeHome, 10.02 vs.\ 18.31 on CDDB-Hard) and raises the accuracy of the last expert alone by 1.50 and 7.36 points, both in every seed, while under the Mahalanobis router it changes ACC by $+$0.20 on OfficeHome and $-$0.16 on CDDB-Hard (Tab.~\ref{tab:supp_dinov2_inherit}).
We compare with UniPrompt-CL, which its authors design and evaluate on DINOv2-B, and with S-iPrompt, the per-domain prompt learner of our diagnosis (Sec.~\xref{sec:diagnosis}); Hash, HiDe-Prompt, CONEC-LoRA, EASE and SOYO were not run on DINOv2.
\input{sec/supp_dinov2}
\input{sec/supp_dinov2_inherit}

%% file: sec/supp_baselines.tex
\begin{table*}[t]
\centering
\caption{\textbf{Training settings in our runs.} All methods learn each domain for 15 epochs at batch size 128. The optimizer and initial learning rate (LR) apply to every trained parameter group and phase unless a cell states otherwise; WD: weight decay; $|\mathcal{Y}|$: number of classes.}
\label{tab:supp_baselines}
\scriptsize
\setlength{\tabcolsep}{3pt}
\renewcommand{\arraystretch}{1.1}
\begin{tabular}{@{}lllll>{\raggedright\arraybackslash}p{4.1cm}>{\raggedright\arraybackslash}p{2.3cm}>{\raggedright\arraybackslash}p{3.9cm}@{}}
\toprule
Method & Optimizer & LR & Schedule & WD & Prompt length / pool / module & Auxiliary phases (epochs) & Other settings \\
\midrule
Joint linear probe & Adam & $10^{-3}$ & cosine & 0 & -- & -- & linear head on the pooled TRAIN domains \\
\midrule
SLCA & SGD & 0.01 & multi-step & $5{\times}10^{-4}$ & fine-tuned backbone & classifier alignment (5) & LR step at epoch 12; backbone LR $\times$0.01; alignment on 256 sampled features per class with logit normalization 0.1; prediction modulo $|\mathcal{Y}|$ over the per-domain heads \\
DUCT & SGD & 0.1 & multi-step$^{\ddagger}$ & $5{\times}10^{-4}$ & fine-tuned backbone & classifier retraining (5; LR $10^{-3}$) & backbone LR $\times$0.01; merge ratio 0.25 (OfficeHome, DomainNet) / 0.5 (CDDB-Hard) and optimal-transport regularization 0.1 (OfficeHome, CDDB-Hard) / 0.01 (DomainNet), as in the released configurations; prediction modulo $|\mathcal{Y}|$ over the per-domain heads; classes without training images left out of the transport; log-domain Sinkhorn solver \\
\midrule
CODA-Prompt & Adam & $10^{-3}$ & cosine & 0 & pool 100, length 8, blocks 1--5 & -- & orthogonality weight 0 (released script) \\
OS-Prompt & Adam & $10^{-3}$ & cosine & 0 & pool 100, length 8, blocks 1--5 & -- & query regularizer 0 \\
OS-Prompt++ & Adam & $10^{-3}$ & cosine & 0 & pool 100, length 8, blocks 1--5 & -- & query regularizer $10^{-4}$ \\
KA-Prompt & Adam & $10^{-3}$ & constant & 0.01 & pool 100 split evenly over domains, length 8, general length 4, top-$k$ 3; general prompt in blocks 1--2, per-domain prompts in blocks 3--5 & exponential-moving-average warm-up (1) & $\tau$ 0.01; auxiliary weight 0.1 \\
CPS-Prompt & Adam & $10^{-3}$ & cosine & 0 & pool 100, length 8, blocks 1--5 & -- & reduction 0.5; critical-score sampling; phase ratio 0.8; temperature 1.0; gradient clipping 1.0 \\
UniPrompt-CL & Adam & $10^{-3}$ & cosine & 0 & pool 500, 100 rows added per domain (earlier rows frozen), 5 prompted blocks & -- & selection-loss weight $10^{-4}$ (released training code) \\
\midrule
S-iPrompt & Adam & $10^{-3}$ & cosine & $2{\times}10^{-4}$ & length 8, one prompt set per domain & -- & 5 $k$-means keys per domain; image-only \\
HiDe-Prompt & Adam & $10^{-3}$ & cosine & 0 & one prompt per domain, length 8 (4 key + 4 value), blocks 1--5 & task-identity head (20), classifier alignment (30) & contrastive weight 0.5; prompt momentum 0.01; gradient clipping 1.0; prediction modulo $|\mathcal{Y}|$ over the (domain, class) outputs \\
EASE & SGD & 0.05 & cosine & $5{\times}10^{-3}$ & one adapter of width 64 per domain (released configuration) & -- & prototype reweighting, $\alpha$ 0.1; each domain is a task of $|\mathcal{Y}|$ classes, prediction modulo $|\mathcal{Y}|$ \\
SOYO & Adam & $10^{-3}$ & cosine & $2{\times}10^{-4}$ & length 8; adapter width 16 & router (30) & 2 Gaussian-mixture components \\
Hash & Adam & $10^{-3}$ & cosine & 0 & pool 100, length 8 (4 key + 4 value), general prompt 5 key + 5 value, top-$k$ 2 with softmax weights, blocks 1--5 & task-identity head (50), classifier alignment (30) & gate activation sigmoid, general prompt and prompted blocks as in the released code; prompt momentum 0.01; gradient clipping 1.0; prediction modulo $|\mathcal{Y}|$ over the (domain, class) outputs \\
CONEC-LoRA & SGD & 0.02 & cosine & $2{\times}10^{-4}$ & LoRA rank 8 on query and value, shared in blocks 1--6 and per domain in blocks 7--12 & router (15; LR $2{\times}10^{-3}$) & router on all 12 blocks; 2 Gaussian-mixture components \\
\midrule
Lamarck (Ours) & Adam & $10^{-3}$ & cosine & 0 & pool 100, length 1, blocks 1--5; adapters of rank $r{=}32$ after blocks 6--12 & -- & MLP readout; training-free Mahalanobis router \\
\bottomrule
\multicolumn{8}{@{}l}{$^{\ddagger}$DUCT's released configuration pairs its 15 epochs with milestones at epochs 60, 100 and 140, so the LR stays constant, in the release as in our runs.}
\end{tabular}
\end{table*}

%% file: sec/supp_perdomain_all.tex
\begin{table}[t]
\centering
\caption{\textbf{Own-domain accuracy of the analyzed expert banks.} As Tab.~\xref{tab:perdomain}, with S-iPrompt; TEST, mean$\pm$s.d.\ over five seeds. WFIR: WhichFaceIsReal.}
\label{tab:supp_perdomain}
\scriptsize
\setlength{\tabcolsep}{4.5pt}
\renewcommand{\arraystretch}{1.1}
\begin{adjustbox}{max width=\linewidth}
\begin{tabular}{@{}llcccc@{}}
\toprule
 & & \multicolumn{4}{c}{Own-domain accuracy (\%) $\uparrow$} \\
\cmidrule(l){3-6}
 & Domain & S-iPrompt & SOYO & Lamarck, prompts only & \textbf{Lamarck (Ours)} \\
\midrule
\multirow{6}{*}{\rotatebox[origin=c]{90}{DomainNet}} & Clipart & $77.86{\scriptstyle\pm0.13}$ & $80.85{\scriptstyle\pm0.25}$ & $77.89{\scriptstyle\pm0.15}$ & $80.34{\scriptstyle\pm0.31}$ \\
 & Infograph & $43.54{\scriptstyle\pm0.19}$ & $46.76{\scriptstyle\pm0.15}$ & $43.54{\scriptstyle\pm0.19}$ & $46.11{\scriptstyle\pm0.48}$ \\
 & Painting & $74.20{\scriptstyle\pm0.10}$ & $76.08{\scriptstyle\pm0.18}$ & $74.54{\scriptstyle\pm0.23}$ & $75.55{\scriptstyle\pm0.12}$ \\
 & QuickDraw & $63.86{\scriptstyle\pm0.22}$ & $72.29{\scriptstyle\pm0.18}$ & $64.12{\scriptstyle\pm0.11}$ & $69.79{\scriptstyle\pm0.09}$ \\
 & Real & $86.88{\scriptstyle\pm0.04}$ & $87.41{\scriptstyle\pm0.04}$ & $87.21{\scriptstyle\pm0.19}$ & $87.17{\scriptstyle\pm0.09}$ \\
 & Sketch & $69.22{\scriptstyle\pm0.20}$ & $73.47{\scriptstyle\pm0.14}$ & $71.00{\scriptstyle\pm0.16}$ & $72.64{\scriptstyle\pm0.17}$ \\
\midrule
\multirow{4}{*}{\rotatebox[origin=c]{90}{OfficeHome}} & Art & $83.88{\scriptstyle\pm0.89}$ & $81.55{\scriptstyle\pm1.34}$ & $83.25{\scriptstyle\pm1.25}$ & $83.79{\scriptstyle\pm1.48}$ \\
 & Clipart & $82.38{\scriptstyle\pm0.67}$ & $81.73{\scriptstyle\pm1.24}$ & $83.12{\scriptstyle\pm0.49}$ & $84.05{\scriptstyle\pm0.47}$ \\
 & Product & $96.20{\scriptstyle\pm0.30}$ & $95.37{\scriptstyle\pm0.32}$ & $95.98{\scriptstyle\pm0.20}$ & $96.12{\scriptstyle\pm0.45}$ \\
 & Real World & $92.77{\scriptstyle\pm0.49}$ & $91.40{\scriptstyle\pm0.46}$ & $93.31{\scriptstyle\pm0.46}$ & $93.19{\scriptstyle\pm0.42}$ \\
\midrule
\multirow{5}{*}{\rotatebox[origin=c]{90}{CDDB-Hard}} & GauGAN & $97.85{\scriptstyle\pm0.38}$ & $99.78{\scriptstyle\pm0.10}$ & $98.08{\scriptstyle\pm0.48}$ & $99.58{\scriptstyle\pm0.11}$ \\
 & BigGAN & $91.75{\scriptstyle\pm0.72}$ & $99.31{\scriptstyle\pm0.41}$ & $94.62{\scriptstyle\pm0.60}$ & $99.31{\scriptstyle\pm0.34}$ \\
 & Wild & $76.95{\scriptstyle\pm1.28}$ & $87.84{\scriptstyle\pm0.64}$ & $87.16{\scriptstyle\pm1.05}$ & $92.29{\scriptstyle\pm0.34}$ \\
 & WFIR & $89.62{\scriptstyle\pm0.95}$ & $72.00{\scriptstyle\pm5.85}$ & $91.12{\scriptstyle\pm1.20}$ & $97.88{\scriptstyle\pm0.71}$ \\
 & SAN & $78.79{\scriptstyle\pm11.93}$ & $81.82{\scriptstyle\pm0.00}$ & $75.76{\scriptstyle\pm4.79}$ & $84.85{\scriptstyle\pm2.14}$ \\
\bottomrule
\end{tabular}
\end{adjustbox}
\end{table}

%% file: sec/supp_retrieval.tex
\begin{table}[t]
\centering
\caption{\textbf{Prompt retrieval is nearly uniform and nearly input-independent.} Retrieval weights of Lamarck's last prompt-only expert (one seed per benchmark) over up to 256 VAL images per domain and 100 visible pool rows (uniform weight 0.01), per prompted block: entropy relative to its maximum, mean largest weight, and the total-variation (TV) distance of an image's weights from the average weights (mean and maximum over images).}
\label{tab:supp_retrieval}
\scriptsize
\setlength{\tabcolsep}{3pt}
\begin{tabular}{@{}lccccc@{}}
\toprule
Benchmark & Block & Entropy/$\log Z$ & Largest weight & TV mean & TV max \\
\midrule
DomainNet & 1 & 99.14\% & 0.0169 & 0.0000 & 0.0000 \\
 & 2 & 99.01\% & 0.0177 & 0.0008 & 0.0027 \\
 & 3 & 98.47\% & 0.0179 & 0.0039 & 0.0090 \\
 & 4 & 99.56\% & 0.0156 & 0.0081 & 0.0162 \\
 & 5 & 98.96\% & 0.0211 & 0.0075 & 0.0200 \\
\midrule
OfficeHome & 1 & 99.94\% & 0.0117 & 0.0000 & 0.0000 \\
 & 2 & 99.89\% & 0.0124 & 0.0005 & 0.0016 \\
 & 3 & 99.95\% & 0.0114 & 0.0013 & 0.0034 \\
 & 4 & 99.95\% & 0.0118 & 0.0016 & 0.0041 \\
 & 5 & 99.95\% & 0.0122 & 0.0022 & 0.0051 \\
\midrule
CDDB-Hard & 1 & 99.95\% & 0.0121 & 0.0000 & 0.0000 \\
 & 2 & 99.92\% & 0.0129 & 0.0004 & 0.0013 \\
 & 3 & 99.95\% & 0.0117 & 0.0012 & 0.0025 \\
 & 4 & 99.92\% & 0.0140 & 0.0015 & 0.0033 \\
 & 5 & 99.90\% & 0.0135 & 0.0022 & 0.0045 \\
\bottomrule
\end{tabular}
\end{table}

%% file: sec/supp_sir.tex
\begin{table}[t]
\centering
\caption{\textbf{Querying every prompt pool with the router's frozen feature.} Change in OfficeHome accuracy against the prompt-only experts (same seeds) under the prototype router, for softmax temperatures $\tau$ of the retrieval; points, mean$\pm$s.d.\ over five seeds; in parentheses, seeds with a positive change.}
\label{tab:supp_sir}
\scriptsize
\setlength{\tabcolsep}{3pt}
\resizebox{\linewidth}{!}{%
\begin{tabular}{@{}lcccc@{}}
\toprule
$\tau$ & VAL $\Delta$AIA$\uparrow$ & VAL $\Delta$ACC$\uparrow$ & TEST $\Delta$AIA$\uparrow$ & TEST $\Delta$ACC$\uparrow$ \\
\midrule
0.03 & $-0.27\,{\scriptstyle\pm\,0.35}$ (1/5) & $-0.15\,{\scriptstyle\pm\,0.17}$ (1/5) & $-0.14\,{\scriptstyle\pm\,0.37}$ (2/5) & $0.12\,{\scriptstyle\pm\,0.33}$ (4/5) \\
0.1 & $0.01\,{\scriptstyle\pm\,0.28}$ (2/5) & $0.10\,{\scriptstyle\pm\,0.28}$ (3/5) & $-0.02\,{\scriptstyle\pm\,0.41}$ (4/5) & $0.08\,{\scriptstyle\pm\,0.20}$ (3/5) \\
0.3 & $0.12\,{\scriptstyle\pm\,0.25}$ (4/5) & $0.11\,{\scriptstyle\pm\,0.23}$ (2/5) & $-0.09\,{\scriptstyle\pm\,0.44}$ (2/5) & $0.07\,{\scriptstyle\pm\,0.14}$ (4/5) \\
\bottomrule
\end{tabular}}
\end{table}

%% file: sec/supp_table_duct.tex
\begin{table*}[t]
\centering
\caption{\textbf{SLCA and DUCT, which fine-tune the backbone, lead on OfficeHome but trail Lamarck on CDDB-Hard and DomainNet.} Five-seed TEST accuracy (\%, mean$\pm$s.d.) on the splits, domain orders and seeds of Tab.~\xref{tab:system}; settings in Tab.~\ref{tab:supp_baselines}.}
\label{tab:supp_duct}
\scriptsize
\setlength{\tabcolsep}{3pt}
\begin{tabular}{@{}lcccccc@{}}
\toprule
\multirow{2}{*}{Method} & \multicolumn{2}{c}{DomainNet} & \multicolumn{2}{c}{OfficeHome} & \multicolumn{2}{c}{CDDB-Hard} \\
\cmidrule(lr){2-3}\cmidrule(lr){4-5}\cmidrule(l){6-7}
& AIA$\uparrow$ & ACC$\uparrow$ & AIA$\uparrow$ & ACC$\uparrow$ & AIA$\uparrow$ & ACC$\uparrow$ \\
\midrule
SLCA~\cite{zhang2023slca} & $68.71\,{\scriptstyle\pm\,0.05}$ & $67.97\,{\scriptstyle\pm\,0.07}$ & $88.16\,{\scriptstyle\pm\,0.41}$ & $90.14\,{\scriptstyle\pm\,0.46}$ & $95.52\,{\scriptstyle\pm\,0.22}$ & $89.35\,{\scriptstyle\pm\,1.37}$ \\
DUCT~\cite{zhou2025dual} & $63.58\,{\scriptstyle\pm\,0.06}$ & $59.96\,{\scriptstyle\pm\,0.12}$ & $86.71\,{\scriptstyle\pm\,0.11}$ & $88.87\,{\scriptstyle\pm\,0.33}$ & $81.65\,{\scriptstyle\pm\,2.55}$ & $67.58\,{\scriptstyle\pm\,3.86}$ \\
\textbf{Lamarck (Ours)} & $68.89\,{\scriptstyle\pm\,0.20}$ & $69.46\,{\scriptstyle\pm\,0.14}$ & $85.85\,{\scriptstyle\pm\,0.63}$ & $88.47\,{\scriptstyle\pm\,0.25}$ & $97.16\,{\scriptstyle\pm\,0.17}$ & $92.94\,{\scriptstyle\pm\,0.62}$ \\
\bottomrule
\end{tabular}
\end{table*}

%% file: sec/supp_cddb_views.tex
\begin{table*}[t]
\centering
\caption{\textbf{Lamarck is first on CDDB-Hard under every weighting of the five domains.} TEST accuracy after the last domain (\%, mean$\pm$s.d.\ over five seeds), each method with its own inference rule. Equal: domain mean (ACC of Tab.~\xref{tab:system}); Images: all 2{,}140 TEST images; w/o SAN: without the last domain; in parentheses, TEST images per domain. Best and second best among the learners of Tab.~\xref{tab:system} in bold and underlined.}
\label{tab:supp_cddb_views}
\scriptsize
\setlength{\tabcolsep}{2.6pt}
\resizebox{\textwidth}{!}{%
\begin{tabular}{@{}lccccccccc@{}}
\toprule
 & \multicolumn{4}{c}{Aggregate accuracy$\uparrow$} & \multicolumn{5}{c}{Per-domain accuracy$\uparrow$} \\
\cmidrule(lr){2-5}\cmidrule(l){6-10}
Method & Equal & Images & Equal w/o SAN & Images w/o SAN & GauGAN (800) & BigGAN (320) & Wild (827) & WhichFaceIsReal (160) & SAN (33) \\
\midrule
Joint linear probe & $80.63\,{\scriptstyle\pm\,0.34}$ & $82.46\,{\scriptstyle\pm\,0.18}$ & $82.45\,{\scriptstyle\pm\,0.25}$ & $82.60\,{\scriptstyle\pm\,0.19}$ & $94.90\,{\scriptstyle\pm\,0.29}$ & $81.12\,{\scriptstyle\pm\,0.90}$ & $71.29\,{\scriptstyle\pm\,0.52}$ & $82.50\,{\scriptstyle\pm\,1.17}$ & $73.33\,{\scriptstyle\pm\,1.36}$ \\
\midrule
SLCA~\cite{zhang2023slca} (fine-tuned backbone) & $89.35\,{\scriptstyle\pm\,1.37}$ & $88.45\,{\scriptstyle\pm\,1.49}$ & $90.17\,{\scriptstyle\pm\,1.97}$ & $88.49\,{\scriptstyle\pm\,1.54}$ & $88.88\,{\scriptstyle\pm\,3.22}$ & $91.06\,{\scriptstyle\pm\,3.43}$ & $85.88\,{\scriptstyle\pm\,1.21}$ & $94.88\,{\scriptstyle\pm\,2.27}$ & $86.06\,{\scriptstyle\pm\,3.46}$ \\
DUCT~\cite{zhou2025dual} (fine-tuned backbone) & $67.58\,{\scriptstyle\pm\,3.86}$ & $74.55\,{\scriptstyle\pm\,4.87}$ & $73.27\,{\scriptstyle\pm\,4.41}$ & $75.02\,{\scriptstyle\pm\,4.93}$ & $89.95\,{\scriptstyle\pm\,5.47}$ & $87.00\,{\scriptstyle\pm\,5.84}$ & $59.49\,{\scriptstyle\pm\,6.58}$ & $56.62\,{\scriptstyle\pm\,2.88}$ & $44.85\,{\scriptstyle\pm\,4.49}$ \\
\midrule
CODA-Prompt~\cite{smith2023coda} & $72.92\,{\scriptstyle\pm\,2.33}$ & $69.88\,{\scriptstyle\pm\,1.89}$ & $69.48\,{\scriptstyle\pm\,2.39}$ & $69.62\,{\scriptstyle\pm\,1.92}$ & $73.30\,{\scriptstyle\pm\,2.71}$ & $75.94\,{\scriptstyle\pm\,1.53}$ & $64.69\,{\scriptstyle\pm\,1.56}$ & $64.00\,{\scriptstyle\pm\,5.69}$ & $86.67\,{\scriptstyle\pm\,7.30}$ \\
OS-Prompt~\cite{kim2024one} & $72.84\,{\scriptstyle\pm\,2.14}$ & $69.46\,{\scriptstyle\pm\,2.72}$ & $68.93\,{\scriptstyle\pm\,2.38}$ & $69.16\,{\scriptstyle\pm\,2.74}$ & $74.83\,{\scriptstyle\pm\,2.95}$ & $79.06\,{\scriptstyle\pm\,1.34}$ & $61.57\,{\scriptstyle\pm\,4.32}$ & $60.25\,{\scriptstyle\pm\,2.71}$ & $\mathbf{88.48}\,{\scriptstyle\pm\,2.54}$ \\
OS-Prompt++~\cite{kim2024one} & $72.69\,{\scriptstyle\pm\,2.21}$ & $69.57\,{\scriptstyle\pm\,2.99}$ & $69.05\,{\scriptstyle\pm\,2.78}$ & $69.29\,{\scriptstyle\pm\,3.04}$ & $74.60\,{\scriptstyle\pm\,3.47}$ & $79.62\,{\scriptstyle\pm\,1.44}$ & $61.96\,{\scriptstyle\pm\,4.40}$ & $60.00\,{\scriptstyle\pm\,3.22}$ & $\underline{87.27}\,{\scriptstyle\pm\,1.36}$ \\
KA-Prompt~\cite{xu2025kaprompt} & $74.96\,{\scriptstyle\pm\,0.39}$ & $74.06\,{\scriptstyle\pm\,0.56}$ & $74.91\,{\scriptstyle\pm\,0.43}$ & $74.04\,{\scriptstyle\pm\,0.57}$ & $78.20\,{\scriptstyle\pm\,1.10}$ & $69.50\,{\scriptstyle\pm\,1.30}$ & $70.30\,{\scriptstyle\pm\,1.10}$ & $81.62\,{\scriptstyle\pm\,1.30}$ & $75.15\,{\scriptstyle\pm\,1.36}$ \\
CPS-Prompt~\cite{lim2026critical} & $71.45\,{\scriptstyle\pm\,1.47}$ & $68.01\,{\scriptstyle\pm\,1.90}$ & $67.65\,{\scriptstyle\pm\,1.51}$ & $67.72\,{\scriptstyle\pm\,1.91}$ & $76.92\,{\scriptstyle\pm\,1.59}$ & $74.12\,{\scriptstyle\pm\,2.61}$ & $57.41\,{\scriptstyle\pm\,3.28}$ & $62.12\,{\scriptstyle\pm\,1.57}$ & $86.67\,{\scriptstyle\pm\,1.66}$ \\
UniPrompt-CL~\cite{oh2025towards} & $76.14\,{\scriptstyle\pm\,0.89}$ & $76.60\,{\scriptstyle\pm\,0.65}$ & $76.09\,{\scriptstyle\pm\,0.47}$ & $76.60\,{\scriptstyle\pm\,0.63}$ & $78.33\,{\scriptstyle\pm\,1.53}$ & $81.50\,{\scriptstyle\pm\,1.89}$ & $74.27\,{\scriptstyle\pm\,2.24}$ & $70.25\,{\scriptstyle\pm\,3.44}$ & $76.36\,{\scriptstyle\pm\,4.49}$ \\
\midrule
S-iPrompt~\cite{wang2022sprompts} & $85.60\,{\scriptstyle\pm\,1.92}$ & $86.39\,{\scriptstyle\pm\,0.63}$ & $87.15\,{\scriptstyle\pm\,0.34}$ & $86.50\,{\scriptstyle\pm\,0.55}$ & $95.80\,{\scriptstyle\pm\,0.61}$ & $86.31\,{\scriptstyle\pm\,0.26}$ & $77.00\,{\scriptstyle\pm\,1.28}$ & $89.50\,{\scriptstyle\pm\,0.81}$ & $79.39\,{\scriptstyle\pm\,8.41}$ \\
HiDe-Prompt~\cite{wang2023hierarchical} & $86.39\,{\scriptstyle\pm\,0.61}$ & $87.90\,{\scriptstyle\pm\,0.43}$ & $88.60\,{\scriptstyle\pm\,0.48}$ & $88.06\,{\scriptstyle\pm\,0.43}$ & $96.53\,{\scriptstyle\pm\,0.37}$ & $89.38\,{\scriptstyle\pm\,0.91}$ & $79.11\,{\scriptstyle\pm\,0.73}$ & $89.38\,{\scriptstyle\pm\,1.33}$ & $77.58\,{\scriptstyle\pm\,1.66}$ \\
EASE~\cite{zhou2024expandable} & $67.14\,{\scriptstyle\pm\,0.79}$ & $75.38\,{\scriptstyle\pm\,0.51}$ & $72.56\,{\scriptstyle\pm\,0.99}$ & $75.85\,{\scriptstyle\pm\,0.52}$ & $97.60\,{\scriptstyle\pm\,0.46}$ & $72.12\,{\scriptstyle\pm\,1.89}$ & $59.03\,{\scriptstyle\pm\,0.35}$ & $61.50\,{\scriptstyle\pm\,2.01}$ & $45.45\,{\scriptstyle\pm\,0.00}$ \\
SOYO~\cite{wang2025soyo} & $79.24\,{\scriptstyle\pm\,0.95}$ & $\underline{91.21}\,{\scriptstyle\pm\,0.50}$ & $87.53\,{\scriptstyle\pm\,1.10}$ & $\underline{91.92}\,{\scriptstyle\pm\,0.50}$ & $\underline{99.05}\,{\scriptstyle\pm\,0.21}$ & $\underline{98.00}\,{\scriptstyle\pm\,0.36}$ & $\underline{87.84}\,{\scriptstyle\pm\,0.64}$ & $65.25\,{\scriptstyle\pm\,3.99}$ & $46.06\,{\scriptstyle\pm\,1.36}$ \\
Hash~\cite{li2026hash} & $\underline{87.75}\,{\scriptstyle\pm\,0.88}$ & $88.80\,{\scriptstyle\pm\,0.55}$ & $\underline{89.99}\,{\scriptstyle\pm\,0.67}$ & $88.96\,{\scriptstyle\pm\,0.54}$ & $96.22\,{\scriptstyle\pm\,0.68}$ & $91.81\,{\scriptstyle\pm\,1.07}$ & $80.31\,{\scriptstyle\pm\,0.65}$ & $\underline{91.62}\,{\scriptstyle\pm\,1.80}$ & $78.79\,{\scriptstyle\pm\,2.14}$ \\
CONEC-LoRA~\cite{paeedeh2025conec} & $79.96\,{\scriptstyle\pm\,2.34}$ & $86.36\,{\scriptstyle\pm\,4.73}$ & $88.28\,{\scriptstyle\pm\,3.51}$ & $86.99\,{\scriptstyle\pm\,4.84}$ & $98.67\,{\scriptstyle\pm\,0.57}$ & $95.62\,{\scriptstyle\pm\,1.23}$ & $72.45\,{\scriptstyle\pm\,11.36}$ & $86.38\,{\scriptstyle\pm\,3.84}$ & $46.67\,{\scriptstyle\pm\,3.46}$ \\
\midrule
\textbf{Lamarck (Ours)} & $\mathbf{92.94}\,{\scriptstyle\pm\,0.62}$ & $\mathbf{96.01}\,{\scriptstyle\pm\,0.13}$ & $\mathbf{96.93}\,{\scriptstyle\pm\,0.25}$ & $\mathbf{96.31}\,{\scriptstyle\pm\,0.15}$ & $\mathbf{99.42}\,{\scriptstyle\pm\,0.11}$ & $\mathbf{98.12}\,{\scriptstyle\pm\,0.49}$ & $\mathbf{92.29}\,{\scriptstyle\pm\,0.34}$ & $\mathbf{97.88}\,{\scriptstyle\pm\,0.71}$ & $76.97\,{\scriptstyle\pm\,2.71}$ \\
\bottomrule
\end{tabular}}
\end{table*}

%% file: sec/supp_revorder.tex
\begin{table}[t]
\centering
\caption{\textbf{Reversing the CDDB-Hard domain order changes Lamarck's ACC by 0.2 points and keeps its lead over Hash and HiDe-Prompt.} TEST accuracy (\%, mean$\pm$s.d.\ over five seeds) with the domains learned in the order of Tab.~\xref{tab:system} (GauGAN, BigGAN, Wild, WFIR, SAN) and in reverse; each baseline with its own inference rule, both Lamarck rows with the Mahalanobis router. $^{\ast}$AIA is not comparable across the two orders: the first reversed stage is SAN alone (33 TEST images).}
\label{tab:supp_revorder}
\scriptsize
\setlength{\tabcolsep}{2.6pt}
\begin{adjustbox}{max width=\linewidth}
\begin{tabular}{@{}lcccc@{}}
\toprule
 & \multicolumn{2}{c}{Original order} & \multicolumn{2}{c}{Reversed order} \\
\cmidrule(lr){2-3}\cmidrule(l){4-5}
Method & AIA$\uparrow$ & ACC$\uparrow$ & AIA$^{\ast}\uparrow$ & ACC$\uparrow$ \\
\midrule
HiDe-Prompt~\cite{wang2023hierarchical} & $91.14\,{\scriptstyle\pm\,0.34}$ & $86.39\,{\scriptstyle\pm\,0.61}$ & $83.71\,{\scriptstyle\pm\,1.08}$ & $86.62\,{\scriptstyle\pm\,0.56}$ \\
Hash~\cite{li2026hash} & $92.37\,{\scriptstyle\pm\,0.36}$ & $87.75\,{\scriptstyle\pm\,0.88}$ & $86.30\,{\scriptstyle\pm\,0.77}$ & $88.46\,{\scriptstyle\pm\,1.07}$ \\
Lamarck, reset adapters & $97.13\,{\scriptstyle\pm\,0.16}$ & $92.73\,{\scriptstyle\pm\,0.63}$ & $90.18\,{\scriptstyle\pm\,1.82}$ & $93.12\,{\scriptstyle\pm\,0.89}$ \\
\textbf{Lamarck (Ours)} & $97.16\,{\scriptstyle\pm\,0.17}$ & $92.94\,{\scriptstyle\pm\,0.62}$ & $90.12\,{\scriptstyle\pm\,1.44}$ & $93.15\,{\scriptstyle\pm\,0.96}$ \\
\bottomrule
\end{tabular}
\end{adjustbox}
\end{table}

%% file: sec/supp_dn_perdomain.tex
\begin{table}[t]
\centering
\caption{\textbf{On DomainNet, Lamarck is more accurate than EASE and CONEC-LoRA on every domain except Infograph and Painting, where CONEC-LoRA leads.} TEST accuracy on each domain after the last domain (\%, mean$\pm$s.d.\ over five seeds), each method with its own inference rule, Lamarck with the Mahalanobis router; ACC: mean over the domains (Tab.~\xref{tab:system}).}
\label{tab:supp_dn_perdomain}
\scriptsize
\setlength{\tabcolsep}{4.5pt}
\begin{tabular}{@{}lccc@{}}
\toprule
Domain & EASE~\cite{zhou2024expandable} & CONEC-LoRA~\cite{paeedeh2025conec} & \textbf{Lamarck (Ours)} \\
\midrule
Clipart & $69.19\,{\scriptstyle\pm\,0.08}$ & $77.10\,{\scriptstyle\pm\,0.17}$ & $77.58\,{\scriptstyle\pm\,0.27}$ \\
Infograph & $35.84\,{\scriptstyle\pm\,0.07}$ & $44.92\,{\scriptstyle\pm\,0.12}$ & $44.16\,{\scriptstyle\pm\,0.44}$ \\
Painting & $67.52\,{\scriptstyle\pm\,0.04}$ & $73.35\,{\scriptstyle\pm\,0.18}$ & $70.79\,{\scriptstyle\pm\,0.16}$ \\
QuickDraw & $50.32\,{\scriptstyle\pm\,0.09}$ & $60.66\,{\scriptstyle\pm\,0.65}$ & $69.76\,{\scriptstyle\pm\,0.09}$ \\
Real & $82.99\,{\scriptstyle\pm\,0.03}$ & $83.70\,{\scriptstyle\pm\,0.09}$ & $85.00\,{\scriptstyle\pm\,0.04}$ \\
Sketch & $62.33\,{\scriptstyle\pm\,0.02}$ & $69.16\,{\scriptstyle\pm\,0.08}$ & $69.47\,{\scriptstyle\pm\,0.23}$ \\
\midrule
ACC$\uparrow$ & $61.36\,{\scriptstyle\pm\,0.02}$ & $68.15\,{\scriptstyle\pm\,0.13}$ & $69.46\,{\scriptstyle\pm\,0.14}$ \\
\bottomrule
\end{tabular}
\end{table}

%% file: sec/supp_eff.tex
\begin{table*}[t]
\centering
\caption{\textbf{Cost of the timed learners on DomainNet after the last domain.} One V100, FP32, batch size 64; protocol in Sec.~\ref{sec:supp_cost}. Passes: encoder forward passes per test image; Param.: parameters added to the backbone and used at test time. $^{\S}$Without the class statistics (9.1\,GB) that the released code loads with the checkpoint but never reads at test time; training memory includes them. $^{\dagger}$Timed on the final-stage architecture with initial weights.}
\label{tab:supp_eff}
\small
\setlength{\tabcolsep}{5pt}
\begin{tabular}{@{}lcccccccc@{}}
\toprule
\multirow{2}{*}{Method} & \multicolumn{5}{c}{Inference} & \multicolumn{3}{c}{Training step} \\
\cmidrule(lr){2-6}\cmidrule(l){7-9}
& Passes$\downarrow$ & GFLOPs$\downarrow$ & Param.\,(M)$\downarrow$ & img/s$\uparrow$ & Mem.\,(GB)$\downarrow$ & GFLOPs$\downarrow$ & img/s$\uparrow$ & Mem.\,(GB)$\downarrow$ \\
\midrule
CODA-Prompt~\cite{smith2023coda} & 2 & 70.3 & 4.11 & 159.1 & 0.90 & 105.9 & 100.9 & 6.15 \\
OS-Prompt~\cite{kim2024one} & 1 & 35.1 & 3.72 & 314.8 & 0.86 & 70.8 & 146.3 & 6.09 \\
OS-Prompt++~\cite{kim2024one} & 1 & 35.1 & 3.72 & 314.6 & 1.18 & 105.9 & 100.5 & 6.45 \\
KA-Prompt~\cite{xu2025kaprompt} & 2 & 88.1 & 6.76 & 130.0 & 1.45 & 283.0 & 38.2 & 15.68 \\
CPS-Prompt~\cite{lim2026critical} & 2 & 70.3 & 4.11 & 159.7 & 0.97 & 69.7 & 151.2 & 3.09 \\
UniPrompt-CL~\cite{oh2025towards} & 1 & 35.9 & 1.96 & 326.6 & 0.82 & 73.8 & 144.3 & 5.38 \\
\midrule
S-iPrompt~\cite{wang2022sprompts} & 2 & 71.7 & 1.65 & 154.5 & 0.87 & 74.5 & 148.3 & 6.26 \\
HiDe-Prompt~\cite{wang2023hierarchical} & 2 & 70.3 & 5.73 & 158.6 & 1.20$^{\S}$ & 106.1 & 102.2 & 13.87 \\
EASE~\cite{zhou2024expandable}$^{\dagger}$ & 6 & 212.4 & 16.68 & 51.5 & 1.05 & 69.8 & 151.2 & 5.72 \\
SOYO~\cite{wang2025soyo} & 7 & 255.5 & 3.46 & 43.5 & 0.86 & 74.9 & 143.5 & 6.73 \\
Hash~\cite{li2026hash} & 2 & 71.0 & 9.12 & 154.7 & 1.21$^{\S}$ & 73.1 & 142.6 & 14.30 \\
CONEC-LoRA~\cite{paeedeh2025conec}$^{\dagger}$ & 2 & 70.4 & 23.16 & 146.2 & 1.23 & 123.7 & 84.7 & 6.24 \\
\midrule
\textbf{Lamarck (Ours)} & 2 & 71.2 & 9.84 & 147.6 & 1.73 & 74.3 & 141.6 & 6.25 \\
\bottomrule
\end{tabular}
\end{table*}

%% file: sec/supp_rank.tex
\begin{table}[t]
\centering
\caption{\textbf{Accuracy depends little on the rank and placement of the adapters.} Rows $r{=}8$, 32, 64 (post-hoc sweep): Lamarck's adapters in blocks 6--12 at rank $r$ (Lamarck uses $r{=}32$); Blocks 1--12: adapters of rank 16 in all twelve blocks, with Lamarck's prompts and inheritance kept. ACC under the Mahalanobis router on VAL and TEST (\%, mean$\pm$s.d.\ over five seeds) and TEST later-expert cost (Eq.~\xeqref{eq:costs}; five-seed mean); --: not computed.}
\label{tab:supp_rank}
\scriptsize
\setlength{\tabcolsep}{2.6pt}
\begin{tabular}{@{}llccc@{}}
\toprule
Benchmark & Adapters & VAL ACC$\uparrow$ & TEST ACC$\uparrow$ & Later-expert cost$\downarrow$ \\
\midrule
DomainNet & $r{=}8$ & $69.11\,{\scriptstyle\pm\,0.18}$ & $69.18\,{\scriptstyle\pm\,0.04}$ & 31.43 \\
 & $r{=}32$ & $69.21\,{\scriptstyle\pm\,0.11}$ & $69.46\,{\scriptstyle\pm\,0.14}$ & 31.47 \\
 & $r{=}64$ & $69.37\,{\scriptstyle\pm\,0.08}$ & $69.56\,{\scriptstyle\pm\,0.05}$ & 31.34 \\
 & Blocks 1--12 & -- & $69.49\,{\scriptstyle\pm\,0.05}$ & 31.46 \\
\midrule
OfficeHome & $r{=}8$ & $88.95\,{\scriptstyle\pm\,0.18}$ & $88.50\,{\scriptstyle\pm\,0.49}$ & 4.79 \\
 & $r{=}32$ & $89.06\,{\scriptstyle\pm\,0.58}$ & $88.47\,{\scriptstyle\pm\,0.25}$ & 5.04 \\
 & $r{=}64$ & $89.11\,{\scriptstyle\pm\,0.41}$ & $88.43\,{\scriptstyle\pm\,0.20}$ & 5.18 \\
 & Blocks 1--12 & -- & $88.54\,{\scriptstyle\pm\,0.40}$ & 5.41 \\
\midrule
CDDB-Hard & $r{=}8$ & $84.33\,{\scriptstyle\pm\,0.66}$ & $91.93\,{\scriptstyle\pm\,1.35}$ & 12.42 \\
 & $r{=}32$ & $84.70\,{\scriptstyle\pm\,0.60}$ & $92.94\,{\scriptstyle\pm\,0.62}$ & 15.75 \\
 & $r{=}64$ & $85.27\,{\scriptstyle\pm\,0.98}$ & $92.63\,{\scriptstyle\pm\,0.24}$ & 17.61 \\
 & Blocks 1--12 & -- & $93.20\,{\scriptstyle\pm\,0.97}$ & 13.17 \\
\bottomrule
\end{tabular}
\end{table}

%% file: sec/supp_inherit_routing.tex
\begin{table*}[t]
\centering
\caption{\textbf{Inheritance changes ACC little under accurate routing and helps most without a router.} TEST ACC of inherited experts minus that of experts with reset adapters ($-$reset) or of independent experts ($-$indep.); points, mean$\pm$s.d.\ of the per-seed differences over five seeds. Mis.: share (\%) of all TEST images that are not sent to their own domain's expert.}
\label{tab:supp_inherit_routing}
\footnotesize
\setlength{\tabcolsep}{4.5pt}
\begin{tabular}{@{}lccccccccc@{}}
\toprule
 & \multicolumn{3}{c}{DomainNet} & \multicolumn{3}{c}{OfficeHome} & \multicolumn{3}{c}{CDDB-Hard} \\
\cmidrule(lr){2-4}\cmidrule(lr){5-7}\cmidrule(l){8-10}
Routing & Mis.$\downarrow$ & $-$reset & $-$indep. & Mis.$\downarrow$ & $-$reset & $-$indep. & Mis.$\downarrow$ & $-$reset & $-$indep. \\
\midrule
Domain-label routing & 0.0 & ${-}0.41\,{\scriptstyle\pm\,0.12}$ & ${-}0.07\,{\scriptstyle\pm\,0.04}$ & 0.0 & ${-}0.01\,{\scriptstyle\pm\,0.19}$ & ${+}0.23\,{\scriptstyle\pm\,0.20}$ & 0.0 & ${+}0.31\,{\scriptstyle\pm\,0.99}$ & ${+}0.55\,{\scriptstyle\pm\,0.86}$ \\
Mahalanobis router & 11.9 & ${-}0.41\,{\scriptstyle\pm\,0.16}$ & ${+}0.08\,{\scriptstyle\pm\,0.07}$ & 16.2 & ${+}0.15\,{\scriptstyle\pm\,0.15}$ & ${+}0.54\,{\scriptstyle\pm\,0.28}$ & 3.6 & ${+}0.21\,{\scriptstyle\pm\,0.39}$ & ${+}0.58\,{\scriptstyle\pm\,0.75}$ \\
Prototype router & 19.5 & ${-}0.32\,{\scriptstyle\pm\,0.16}$ & ${+}0.33\,{\scriptstyle\pm\,0.06}$ & 36.4 & ${+}0.12\,{\scriptstyle\pm\,0.14}$ & ${+}1.10\,{\scriptstyle\pm\,0.34}$ & 10.5 & ${+}0.77\,{\scriptstyle\pm\,0.47}$ & ${+}1.47\,{\scriptstyle\pm\,0.81}$ \\
Last expert only & 88.2 & ${+}1.32\,{\scriptstyle\pm\,0.61}$ & ${+}6.17\,{\scriptstyle\pm\,0.58}$ & 72.1 & ${+}0.73\,{\scriptstyle\pm\,0.42}$ & ${+}4.80\,{\scriptstyle\pm\,0.66}$ & 98.5 & ${+}7.94\,{\scriptstyle\pm\,3.38}$ & ${+}17.75\,{\scriptstyle\pm\,2.38}$ \\
\bottomrule
\end{tabular}
\end{table*}

%% file: sec/supp_inject.tex
\begin{table*}[t]
\centering
\caption{\textbf{Routing-error injection.} TEST ACC difference for three pairs of expert banks when an extra share $p$ of the TEST images goes from the Mahalanobis routes to a random later, any wrong or earlier expert (50 paired draws per seed); the first two rows give the lines of Fig.~\xref{fig:inject}; points, mean$\pm$s.d.\ over five seeds; in parentheses, seeds with a positive difference. Bottom: ACC with the Mahalanobis router and with the last expert only.}
\label{tab:supp_inject}
\scriptsize
\setlength{\tabcolsep}{2.4pt}
\resizebox{\textwidth}{!}{%
\begin{tabular}{@{}llcccccc@{}}
\toprule
 & & \multicolumn{3}{c}{OfficeHome} & \multicolumn{3}{c}{CDDB-Hard} \\
\cmidrule(lr){3-5}\cmidrule(l){6-8}
Difference & Extra errors to & $p{=}0$ & $p{=}10\%$ & $p{=}30\%$ & $p{=}0$ & $p{=}10\%$ & $p{=}30\%$ \\
\midrule
Lamarck $-$ reset adapters & later experts & $0.15\,{\scriptstyle\pm\,0.15}$ (4/5) & $0.25\,{\scriptstyle\pm\,0.15}$ (5/5) & $0.43\,{\scriptstyle\pm\,0.18}$ (5/5) & $0.21\,{\scriptstyle\pm\,0.39}$ (3/5) & $1.02\,{\scriptstyle\pm\,0.31}$ (5/5) & $2.69\,{\scriptstyle\pm\,0.34}$ (5/5) \\
 & any wrong expert & $0.15\,{\scriptstyle\pm\,0.15}$ (4/5) & $0.21\,{\scriptstyle\pm\,0.15}$ (5/5) & $0.35\,{\scriptstyle\pm\,0.15}$ (5/5) & $0.21\,{\scriptstyle\pm\,0.39}$ (3/5) & $0.74\,{\scriptstyle\pm\,0.35}$ (5/5) & $1.85\,{\scriptstyle\pm\,0.42}$ (5/5) \\
 & earlier experts & $0.15\,{\scriptstyle\pm\,0.15}$ (4/5) & $0.15\,{\scriptstyle\pm\,0.14}$ (4/5) & $0.15\,{\scriptstyle\pm\,0.12}$ (4/5) & $0.21\,{\scriptstyle\pm\,0.39}$ (3/5) & $0.18\,{\scriptstyle\pm\,0.34}$ (4/5) & $0.06\,{\scriptstyle\pm\,0.28}$ (3/5) \\
\midrule
Lamarck $-$ independent experts & later experts & $0.54\,{\scriptstyle\pm\,0.28}$ (5/5) & $1.13\,{\scriptstyle\pm\,0.24}$ (5/5) & $2.32\,{\scriptstyle\pm\,0.19}$ (5/5) & $0.58\,{\scriptstyle\pm\,0.75}$ (4/5) & $2.22\,{\scriptstyle\pm\,0.73}$ (5/5) & $5.58\,{\scriptstyle\pm\,0.69}$ (5/5) \\
 & any wrong expert & $0.54\,{\scriptstyle\pm\,0.28}$ (5/5) & $1.02\,{\scriptstyle\pm\,0.23}$ (5/5) & $1.97\,{\scriptstyle\pm\,0.19}$ (5/5) & $0.58\,{\scriptstyle\pm\,0.75}$ (4/5) & $1.64\,{\scriptstyle\pm\,0.70}$ (5/5) & $3.80\,{\scriptstyle\pm\,0.59}$ (5/5) \\
 & earlier experts & $0.54\,{\scriptstyle\pm\,0.28}$ (5/5) & $0.64\,{\scriptstyle\pm\,0.25}$ (5/5) & $0.79\,{\scriptstyle\pm\,0.22}$ (5/5) & $0.58\,{\scriptstyle\pm\,0.75}$ (4/5) & $0.57\,{\scriptstyle\pm\,0.72}$ (4/5) & $0.49\,{\scriptstyle\pm\,0.63}$ (4/5) \\
\midrule
Prompts only: inherited $-$ independent & later experts & $0.62\,{\scriptstyle\pm\,0.17}$ (5/5) & $1.23\,{\scriptstyle\pm\,0.17}$ (5/5) & $2.39\,{\scriptstyle\pm\,0.20}$ (5/5) & $-0.85\,{\scriptstyle\pm\,0.96}$ (1/5) & $0.80\,{\scriptstyle\pm\,0.90}$ (4/5) & $4.09\,{\scriptstyle\pm\,0.80}$ (5/5) \\
 & any wrong expert & $0.62\,{\scriptstyle\pm\,0.17}$ (5/5) & $1.13\,{\scriptstyle\pm\,0.19}$ (5/5) & $2.08\,{\scriptstyle\pm\,0.22}$ (5/5) & $-0.85\,{\scriptstyle\pm\,0.96}$ (1/5) & $0.36\,{\scriptstyle\pm\,0.81}$ (3/5) & $2.76\,{\scriptstyle\pm\,0.45}$ (5/5) \\
 & earlier experts & $0.62\,{\scriptstyle\pm\,0.17}$ (5/5) & $0.71\,{\scriptstyle\pm\,0.19}$ (5/5) & $0.84\,{\scriptstyle\pm\,0.25}$ (5/5) & $-0.85\,{\scriptstyle\pm\,0.96}$ (1/5) & $-0.68\,{\scriptstyle\pm\,0.87}$ (1/5) & $-0.46\,{\scriptstyle\pm\,0.55}$ (1/5) \\
\midrule
Expert bank &  & Mahalanobis router & Last expert only &  & Mahalanobis router & Last expert only &  \\
\midrule
Lamarck, independent &  & $87.93\,{\scriptstyle\pm\,0.26}$ & $80.79\,{\scriptstyle\pm\,0.67}$ &  & $92.35\,{\scriptstyle\pm\,0.61}$ & $59.31\,{\scriptstyle\pm\,1.28}$ &  \\
Lamarck, reset adapters &  & $88.32\,{\scriptstyle\pm\,0.16}$ & $84.85\,{\scriptstyle\pm\,0.49}$ &  & $92.73\,{\scriptstyle\pm\,0.63}$ & $69.12\,{\scriptstyle\pm\,2.22}$ &  \\
\textbf{Lamarck (Ours)} &  & $88.47\,{\scriptstyle\pm\,0.25}$ & $85.59\,{\scriptstyle\pm\,0.14}$ &  & $92.94\,{\scriptstyle\pm\,0.62}$ & $77.06\,{\scriptstyle\pm\,2.15}$ &  \\
\bottomrule
\end{tabular}}
\end{table*}

%% file: sec/supp_kshot.tex
\begin{table*}[t]
\centering
\caption{\textbf{Lamarck's ACC difference from reset adapters and from independent experts rises as routers send more TEST images to later experts.} Top: values behind the markers of Fig.~\xref{fig:inject}. $M$, $F$: share (\%) of TEST images routed to another or to a later expert, mean over domains. $-$reset, $-$indep.: TEST ACC of Lamarck minus that with reset adapters or of independent experts; points, mean$\pm$s.d.\ over five seeds, in parentheses seeds with a positive difference. Own, Later: parts of $-$reset from images routed to their own or to a later expert (the rest is from earlier experts). $k$-shot: the Mahalanobis router fitted on $k$ TRAIN images per domain, averaged over 20 draws per seed; Euclidean, cosine: nearest TRAIN mean. Slope: per-seed least-squares slope of the difference on $F$ over the $k$-shot draws and the Mahalanobis router (points per point of $F$). Mis.\ in Tab.~\ref{tab:supp_inherit_routing} is $M$ weighted by images instead of domains. Bottom: DomainNet.}
\label{tab:supp_kshot}
\scriptsize
\setlength{\tabcolsep}{2.2pt}
\resizebox{\textwidth}{!}{%
\begin{tabular}{@{}llrrcrrcrrcrrc@{}}
\toprule
 & & \multicolumn{6}{c}{OfficeHome} & \multicolumn{6}{c}{CDDB-Hard} \\
\cmidrule(lr){3-8}\cmidrule(l){9-14}
Router & $k$ & $M$ & $F$ & $-$reset & Own & Later & $-$indep. & $M$ & $F$ & $-$reset & Own & Later & $-$indep. \\
\midrule
Mahalanobis & all & 17.47 & 9.65 & ${+}0.15\,{\scriptstyle\pm\,0.15}$ (4/5) & ${+}0.04$ & ${+}0.06$ & ${+}0.54\,{\scriptstyle\pm\,0.28}$ (5/5) & 6.45 & 1.28 & ${+}0.21\,{\scriptstyle\pm\,0.39}$ (3/5) & ${+}0.32$ & ${+}0.05$ & ${+}0.58\,{\scriptstyle\pm\,0.75}$ (4/5) \\
$k$-shot & 256 & 35.88 & 18.72 & ${+}0.18\,{\scriptstyle\pm\,0.10}$ (5/5) & $0.00$ & ${+}0.13$ & ${+}1.27\,{\scriptstyle\pm\,0.12}$ (5/5) & 13.42 & 6.44 & ${+}0.70\,{\scriptstyle\pm\,0.46}$ (5/5) & ${+}0.29$ & ${+}0.53$ & ${+}1.18\,{\scriptstyle\pm\,0.77}$ (4/5) \\
 & 128 & 39.59 & 20.56 & ${+}0.29\,{\scriptstyle\pm\,0.11}$ (5/5) & ${-}0.01$ & ${+}0.23$ & ${+}1.46\,{\scriptstyle\pm\,0.15}$ (5/5) & 18.61 & 9.36 & ${+}1.09\,{\scriptstyle\pm\,0.39}$ (5/5) & ${+}0.29$ & ${+}0.91$ & ${+}1.88\,{\scriptstyle\pm\,0.70}$ (5/5) \\
 & 64 & 36.93 & 19.82 & ${+}0.30\,{\scriptstyle\pm\,0.15}$ (5/5) & $0.00$ & ${+}0.23$ & ${+}1.20\,{\scriptstyle\pm\,0.13}$ (5/5) & 15.80 & 7.27 & ${+}0.81\,{\scriptstyle\pm\,0.37}$ (5/5) & ${+}0.32$ & ${+}0.63$ & ${+}1.49\,{\scriptstyle\pm\,0.56}$ (5/5) \\
 & 32 & 40.04 & 21.73 & ${+}0.29\,{\scriptstyle\pm\,0.13}$ (5/5) & ${+}0.01$ & ${+}0.22$ & ${+}1.25\,{\scriptstyle\pm\,0.17}$ (5/5) & 17.08 & 7.88 & ${+}0.90\,{\scriptstyle\pm\,0.36}$ (5/5) & ${+}0.31$ & ${+}0.71$ & ${+}1.74\,{\scriptstyle\pm\,0.65}$ (5/5) \\
 & 16 & 46.07 & 24.28 & ${+}0.25\,{\scriptstyle\pm\,0.14}$ (5/5) & ${+}0.01$ & ${+}0.17$ & ${+}1.49\,{\scriptstyle\pm\,0.12}$ (5/5) & 20.46 & 9.86 & ${+}1.24\,{\scriptstyle\pm\,0.34}$ (5/5) & ${+}0.36$ & ${+}0.99$ & ${+}2.34\,{\scriptstyle\pm\,0.44}$ (5/5) \\
 & 8 & 52.06 & 26.90 & ${+}0.38\,{\scriptstyle\pm\,0.15}$ (5/5) & ${+}0.01$ & ${+}0.27$ & ${+}1.97\,{\scriptstyle\pm\,0.16}$ (5/5) & 25.34 & 12.63 & ${+}1.71\,{\scriptstyle\pm\,0.48}$ (5/5) & ${+}0.33$ & ${+}1.40$ & ${+}3.20\,{\scriptstyle\pm\,0.53}$ (5/5) \\
 & 4 & 58.59 & 29.80 & ${+}0.40\,{\scriptstyle\pm\,0.10}$ (5/5) & ${+}0.01$ & ${+}0.31$ & ${+}2.46\,{\scriptstyle\pm\,0.41}$ (5/5) & 30.18 & 15.70 & ${+}1.94\,{\scriptstyle\pm\,0.62}$ (5/5) & ${+}0.30$ & ${+}1.67$ & ${+}3.75\,{\scriptstyle\pm\,0.89}$ (5/5) \\
 & 2 & 63.07 & 32.02 & ${+}0.49\,{\scriptstyle\pm\,0.14}$ (5/5) & ${+}0.01$ & ${+}0.39$ & ${+}3.07\,{\scriptstyle\pm\,0.25}$ (5/5) & 37.47 & 18.37 & ${+}2.32\,{\scriptstyle\pm\,0.78}$ (5/5) & ${+}0.27$ & ${+}2.06$ & ${+}4.42\,{\scriptstyle\pm\,0.55}$ (5/5) \\
\midrule
Prototype & all & 37.27 & 19.31 & ${+}0.12\,{\scriptstyle\pm\,0.14}$ (4/5) & $0.00$ & ${+}0.10$ & ${+}1.10\,{\scriptstyle\pm\,0.34}$ (5/5) & 14.35 & 5.09 & ${+}0.77\,{\scriptstyle\pm\,0.47}$ (5/5) & ${+}0.32$ & ${+}0.51$ & ${+}1.47\,{\scriptstyle\pm\,0.81}$ (5/5) \\
Euclidean & all & 36.56 & 16.90 & ${+}0.17\,{\scriptstyle\pm\,0.17}$ (4/5) & ${+}0.01$ & ${+}0.06$ & ${+}0.80\,{\scriptstyle\pm\,0.21}$ (5/5) & 18.54 & 7.65 & ${+}1.07\,{\scriptstyle\pm\,0.30}$ (5/5) & ${+}0.42$ & ${+}0.84$ & ${+}2.26\,{\scriptstyle\pm\,0.44}$ (5/5) \\
Cosine & all & 35.79 & 16.91 & ${+}0.27\,{\scriptstyle\pm\,0.13}$ (5/5) & ${-}0.03$ & ${+}0.18$ & ${+}0.93\,{\scriptstyle\pm\,0.32}$ (5/5) & 19.38 & 8.84 & ${+}1.15\,{\scriptstyle\pm\,0.71}$ (5/5) & ${+}0.42$ & ${+}1.02$ & ${+}2.41\,{\scriptstyle\pm\,0.66}$ (5/5) \\
\midrule
Last expert only & -- & 75.00 & 75.00 & ${+}0.73\,{\scriptstyle\pm\,0.42}$ (5/5) & ${+}0.01$ & ${+}0.72$ & ${+}4.80\,{\scriptstyle\pm\,0.66}$ (5/5) & 80.00 & 80.00 & ${+}7.94\,{\scriptstyle\pm\,3.38}$ (5/5) & ${+}0.24$ & ${+}7.70$ & ${+}17.75\,{\scriptstyle\pm\,2.38}$ (5/5) \\
\midrule
Slope on $F$ &  &  &  & $0.016\,{\scriptstyle\pm\,0.006}$ &  &  & $0.108\,{\scriptstyle\pm\,0.018}$ &  &  & $0.129\,{\scriptstyle\pm\,0.050}$ &  &  & $0.259\,{\scriptstyle\pm\,0.049}$ \\
\midrule
 & & \multicolumn{6}{c}{DomainNet} &  &  &  &  &  & \\
\cmidrule(lr){3-8}
Router & $k$ & $M$ & $F$ & $-$reset & Own & Later & $-$indep. &  &  &  &  &  &  \\
\midrule
Mahalanobis & all & 15.43 & 9.18 & ${-}0.41\,{\scriptstyle\pm\,0.16}$ (0/5) & ${-}0.38$ & ${-}0.02$ & ${+}0.08\,{\scriptstyle\pm\,0.07}$ (4/5) &  &  &  &  &  &  \\
$k$-shot & 256 & 24.70 & 12.07 & ${-}0.30\,{\scriptstyle\pm\,0.16}$ (0/5) & ${-}0.32$ & ${+}0.04$ & ${+}0.44\,{\scriptstyle\pm\,0.09}$ (5/5) &  &  &  &  &  &  \\
 & 128 & 34.86 & 17.12 & ${-}0.13\,{\scriptstyle\pm\,0.17}$ (2/5) & ${-}0.28$ & ${+}0.14$ & ${+}0.97\,{\scriptstyle\pm\,0.08}$ (5/5) &  &  &  &  &  &  \\
 & 64 & 29.97 & 14.81 & ${-}0.23\,{\scriptstyle\pm\,0.16}$ (0/5) & ${-}0.30$ & ${+}0.07$ & ${+}0.67\,{\scriptstyle\pm\,0.07}$ (5/5) &  &  &  &  &  &  \\
 & 32 & 27.64 & 13.59 & ${-}0.27\,{\scriptstyle\pm\,0.15}$ (0/5) & ${-}0.31$ & ${+}0.04$ & ${+}0.49\,{\scriptstyle\pm\,0.08}$ (5/5) &  &  &  &  &  &  \\
 & 16 & 28.40 & 14.23 & ${-}0.28\,{\scriptstyle\pm\,0.15}$ (0/5) & ${-}0.31$ & ${+}0.04$ & ${+}0.51\,{\scriptstyle\pm\,0.08}$ (5/5) &  &  &  &  &  &  \\
 & 8 & 32.04 & 15.56 & ${-}0.23\,{\scriptstyle\pm\,0.15}$ (0/5) & ${-}0.29$ & ${+}0.06$ & ${+}0.67\,{\scriptstyle\pm\,0.09}$ (5/5) &  &  &  &  &  &  \\
 & 4 & 37.24 & 17.82 & ${-}0.15\,{\scriptstyle\pm\,0.15}$ (0/5) & ${-}0.27$ & ${+}0.11$ & ${+}0.96\,{\scriptstyle\pm\,0.11}$ (5/5) &  &  &  &  &  &  \\
 & 2 & 45.97 & 20.70 & ${+}0.13\,{\scriptstyle\pm\,0.17}$ (4/5) & ${-}0.23$ & ${+}0.26$ & ${+}1.67\,{\scriptstyle\pm\,0.12}$ (5/5) &  &  &  &  &  &  \\
\midrule
Prototype & all & 22.58 & 11.33 & ${-}0.32\,{\scriptstyle\pm\,0.16}$ (0/5) & ${-}0.33$ & ${+}0.03$ & ${+}0.33\,{\scriptstyle\pm\,0.06}$ (5/5) &  &  &  &  &  &  \\
Euclidean & all & 21.77 & 12.01 & ${-}0.36\,{\scriptstyle\pm\,0.16}$ (0/5) & ${-}0.33$ & ${-}0.01$ & ${+}0.22\,{\scriptstyle\pm\,0.06}$ (5/5) &  &  &  &  &  &  \\
Cosine & all & 21.61 & 11.26 & ${-}0.35\,{\scriptstyle\pm\,0.16}$ (0/5) & ${-}0.32$ & ${-}0.01$ & ${+}0.25\,{\scriptstyle\pm\,0.08}$ (5/5) &  &  &  &  &  &  \\
\midrule
Last expert only & -- & 83.33 & 83.33 & ${+}1.32\,{\scriptstyle\pm\,0.61}$ (5/5) & ${-}0.15$ & ${+}1.48$ & ${+}6.17\,{\scriptstyle\pm\,0.58}$ (5/5) &  &  &  &  &  &  \\
\midrule
Slope on $F$ &  &  &  & $0.031\,{\scriptstyle\pm\,0.006}$ &  &  & $0.105\,{\scriptstyle\pm\,0.004}$ &  &  &  &  &  &  \\
\bottomrule
\end{tabular}}
\end{table*}

%% file: sec/supp_revorder_inherit.tex
\begin{table*}[t]
\centering
\caption{\textbf{In both CDDB-Hard domain orders, inheritance lowers the later-expert cost in every seed and adds the most ACC when the last expert classifies every image.} Lamarck and Lamarck with reset adapters in the original and the reversed domain order (Tab.~\ref{tab:supp_revorder}); $\Delta$: Lamarck minus reset adapters, paired by seed, in parentheses seeds with a positive difference, so 0/5 on a cost row means that Lamarck's cost is lower in every seed. Later- and earlier-expert cost as in Eq.~\xeqref{eq:costs}; TEST, mean$\pm$s.d.\ over five seeds.}
\label{tab:supp_revorder_inherit}
\scriptsize
\setlength{\tabcolsep}{3.5pt}
\begin{tabular}{@{}lcccccc@{}}
\toprule
 & \multicolumn{3}{c}{Original order (GauGAN to SAN)} & \multicolumn{3}{c}{Reversed order (SAN to GauGAN)} \\
\cmidrule(lr){2-4}\cmidrule(l){5-7}
Metric & Reset adapters & \textbf{Lamarck (Ours)} & $\Delta$ & Reset adapters & \textbf{Lamarck (Ours)} & $\Delta$ \\
\midrule
ACC, Mahalanobis router & $92.73\,{\scriptstyle\pm\,0.63}$ & $92.94\,{\scriptstyle\pm\,0.62}$ & ${+}0.21\,{\scriptstyle\pm\,0.39}$ (3/5) & $93.12\,{\scriptstyle\pm\,0.89}$ & $93.15\,{\scriptstyle\pm\,0.96}$ & ${+}0.03\,{\scriptstyle\pm\,0.26}$ (2/5) \\
ACC, prototype router & $90.84\,{\scriptstyle\pm\,0.71}$ & $91.61\,{\scriptstyle\pm\,0.73}$ & ${+}0.77\,{\scriptstyle\pm\,0.47}$ (5/5) & $90.96\,{\scriptstyle\pm\,0.87}$ & $91.30\,{\scriptstyle\pm\,1.23}$ & ${+}0.34\,{\scriptstyle\pm\,0.47}$ (3/5) \\
ACC, last expert only & $69.12\,{\scriptstyle\pm\,2.22}$ & $77.06\,{\scriptstyle\pm\,2.15}$ & ${+}7.94\,{\scriptstyle\pm\,3.38}$ (5/5) & $66.93\,{\scriptstyle\pm\,1.00}$ & $70.22\,{\scriptstyle\pm\,2.21}$ & ${+}3.29\,{\scriptstyle\pm\,1.81}$ (5/5) \\
DRA & $94.47\,{\scriptstyle\pm\,0.87}$ & $94.78\,{\scriptstyle\pm\,0.53}$ & ${+}0.31\,{\scriptstyle\pm\,0.99}$ (2/5) & $94.91\,{\scriptstyle\pm\,0.71}$ & $94.81\,{\scriptstyle\pm\,0.50}$ & ${-}0.10\,{\scriptstyle\pm\,0.25}$ (1/5) \\
Later-expert accuracy & $70.06\,{\scriptstyle\pm\,1.41}$ & $82.12\,{\scriptstyle\pm\,0.87}$ & ${+}12.06\,{\scriptstyle\pm\,1.43}$ (5/5) & $57.64\,{\scriptstyle\pm\,1.30}$ & $63.94\,{\scriptstyle\pm\,1.05}$ & ${+}6.30\,{\scriptstyle\pm\,1.38}$ (5/5) \\
Later-expert cost$\downarrow$ & $27.81\,{\scriptstyle\pm\,1.42}$ & $15.75\,{\scriptstyle\pm\,0.84}$ & ${-}12.06\,{\scriptstyle\pm\,1.62}$ (0/5) & $34.25\,{\scriptstyle\pm\,1.10}$ & $27.85\,{\scriptstyle\pm\,1.43}$ & ${-}6.40\,{\scriptstyle\pm\,1.15}$ (0/5) \\
Earlier-expert cost$\downarrow$ & $37.84\,{\scriptstyle\pm\,1.49}$ & $38.93\,{\scriptstyle\pm\,1.46}$ & ${+}1.08\,{\scriptstyle\pm\,2.13}$ (3/5) & $36.66\,{\scriptstyle\pm\,0.71}$ & $37.49\,{\scriptstyle\pm\,1.19}$ & ${+}0.83\,{\scriptstyle\pm\,0.79}$ (5/5) \\
\bottomrule
\end{tabular}
\end{table*}

%% file: sec/supp_revorder_pairs.tex
\begin{table}[t]
\centering
\caption{\textbf{On the same domain pairs, an expert trained after the image's domain costs less than one trained before it, and inheritance enlarges this drop 2.5 to 3.1 times.} Cost (Eq.~\xeqref{eq:costs}) of the ten CDDB-Hard pairs (image domain, expert domain) in which the expert's domain follows the image's domain in the original order (U) and of the ten converse pairs (L), with the expert trained after the image's domain (original order for U, reversed order for L) or before it. Drop: before minus after, paired by seed, in parentheses seeds with a positive drop; TEST, mean$\pm$s.d.\ over five seeds.}
\label{tab:supp_revorder_pairs}
\scriptsize
\setlength{\tabcolsep}{3pt}
\begin{adjustbox}{max width=\linewidth}
\begin{tabular}{@{}llccc@{}}
\toprule
 & & \multicolumn{2}{c}{Cost, expert trained} & \\
\cmidrule(lr){3-4}
Pairs & Expert bank & after$\downarrow$ & before$\downarrow$ & Drop$\uparrow$ \\
\midrule
U & Reset adapters & $27.81\,{\scriptstyle\pm\,1.42}$ & $36.66\,{\scriptstyle\pm\,0.71}$ & ${+}8.85\,{\scriptstyle\pm\,1.43}$ (5/5) \\
 & \textbf{Lamarck (Ours)} & $15.75\,{\scriptstyle\pm\,0.84}$ & $37.49\,{\scriptstyle\pm\,1.19}$ & ${+}21.74\,{\scriptstyle\pm\,1.73}$ (5/5) \\
\midrule
L & Reset adapters & $34.25\,{\scriptstyle\pm\,1.10}$ & $37.84\,{\scriptstyle\pm\,1.49}$ & ${+}3.60\,{\scriptstyle\pm\,1.91}$ (5/5) \\
 & \textbf{Lamarck (Ours)} & $27.85\,{\scriptstyle\pm\,1.43}$ & $38.93\,{\scriptstyle\pm\,1.46}$ & ${+}11.08\,{\scriptstyle\pm\,1.44}$ (5/5) \\
\bottomrule
\end{tabular}
\end{adjustbox}
\end{table}

%% file: sec/supp_dinov2.tex
\begin{table}[t]
\centering
\caption{\textbf{On DINOv2 ViT-B/14, the adapters raise AIA and ACC over Lamarck's prompt-only experts, and Lamarck is above S-iPrompt and UniPrompt-CL.} TEST AIA and ACC (\%), mean$\pm$s.d.\ over five seeds, with the protocol and configuration of ViT-B/16 and no tuning on DINOv2. Lamarck rows use the Mahalanobis router; S-iPrompt and UniPrompt-CL predict as in Tab.~\xref{tab:system}. Inherited and reset adapters are within 0.2 ACC points of each other (Tab.~\ref{tab:supp_dinov2_inherit}).}
\label{tab:supp_dinov2}
\scriptsize
\setlength{\tabcolsep}{3pt}
\begin{adjustbox}{max width=\linewidth}
\begin{tabular}{@{}lcccc@{}}
\toprule
 & \multicolumn{2}{c}{OfficeHome} & \multicolumn{2}{c}{CDDB-Hard} \\
\cmidrule(lr){2-3}\cmidrule(l){4-5}
Method & AIA$\uparrow$ & ACC$\uparrow$ & AIA$\uparrow$ & ACC$\uparrow$ \\
\midrule
S-iPrompt & $85.52\,{\scriptstyle\pm\,0.77}$ & $87.55\,{\scriptstyle\pm\,0.25}$ & $92.39\,{\scriptstyle\pm\,0.28}$ & $86.28\,{\scriptstyle\pm\,0.60}$ \\
UniPrompt-CL & $83.81\,{\scriptstyle\pm\,0.44}$ & $85.69\,{\scriptstyle\pm\,0.19}$ & $91.65\,{\scriptstyle\pm\,0.31}$ & $80.89\,{\scriptstyle\pm\,2.42}$ \\
\midrule
Lamarck, prompts only & $85.81\,{\scriptstyle\pm\,0.81}$ & $88.56\,{\scriptstyle\pm\,0.21}$ & $95.33\,{\scriptstyle\pm\,0.32}$ & $91.16\,{\scriptstyle\pm\,1.11}$ \\
Lamarck, reset adapters & $87.65\,{\scriptstyle\pm\,0.55}$ & $89.15\,{\scriptstyle\pm\,0.40}$ & $97.82\,{\scriptstyle\pm\,0.20}$ & $94.17\,{\scriptstyle\pm\,0.98}$ \\
\textbf{Lamarck (Ours)} & $87.57\,{\scriptstyle\pm\,0.13}$ & $89.34\,{\scriptstyle\pm\,0.09}$ & $97.68\,{\scriptstyle\pm\,0.17}$ & $94.01\,{\scriptstyle\pm\,0.99}$ \\
\bottomrule
\end{tabular}
\end{adjustbox}
\end{table}

%% file: sec/supp_dinov2_inherit.tex
\begin{table*}[t]
\centering
\caption{\textbf{On DINOv2 ViT-B/14, inheritance changes routed ACC by at most 0.2 points, raises ACC with the last expert only and lowers the later-expert cost on both datasets.} TEST ACC (\%) of Lamarck (inherited adapters) and of Lamarck with reset adapters, and Lamarck $-$ reset (points, mean$\pm$s.d.\ of the per-seed differences); last row: later-expert cost (Eq.~\xeqref{eq:costs}). Five seeds. Mis.: share (\%) of all TEST images that are not sent to their own domain's expert, equal in both arms because the routers read the unprompted feature.}
\label{tab:supp_dinov2_inherit}
\scriptsize
\setlength{\tabcolsep}{3pt}
\begin{adjustbox}{max width=\textwidth}
\begin{tabular}{@{}lcccccccc@{}}
\toprule
 & \multicolumn{4}{c}{OfficeHome} & \multicolumn{4}{c}{CDDB-Hard} \\
\cmidrule(lr){2-5}\cmidrule(l){6-9}
Routing & Mis.$\downarrow$ & Lamarck & Reset & $-$reset & Mis.$\downarrow$ & Lamarck & Reset & $-$reset \\
\midrule
Domain-label routing & 0.0 & $90.18\,{\scriptstyle\pm\,0.26}$ & $90.27\,{\scriptstyle\pm\,0.34}$ & ${-}0.08\,{\scriptstyle\pm\,0.23}$ & 0.0 & $94.97\,{\scriptstyle\pm\,0.29}$ & $95.08\,{\scriptstyle\pm\,0.36}$ & ${-}0.11\,{\scriptstyle\pm\,0.41}$ \\
Mahalanobis router & 20.3 & $89.34\,{\scriptstyle\pm\,0.09}$ & $89.15\,{\scriptstyle\pm\,0.40}$ & ${+}0.20\,{\scriptstyle\pm\,0.39}$ & 6.1 & $94.01\,{\scriptstyle\pm\,0.99}$ & $94.17\,{\scriptstyle\pm\,0.98}$ & ${-}0.16\,{\scriptstyle\pm\,0.42}$ \\
Prototype router & 35.2 & $88.61\,{\scriptstyle\pm\,0.30}$ & $88.56\,{\scriptstyle\pm\,0.35}$ & ${+}0.06\,{\scriptstyle\pm\,0.27}$ & 11.3 & $92.64\,{\scriptstyle\pm\,0.65}$ & $92.47\,{\scriptstyle\pm\,0.62}$ & ${+}0.17\,{\scriptstyle\pm\,0.47}$ \\
Last expert only & 72.1 & $86.04\,{\scriptstyle\pm\,0.53}$ & $84.53\,{\scriptstyle\pm\,0.35}$ & ${+}1.50\,{\scriptstyle\pm\,0.58}$ & 98.5 & $84.60\,{\scriptstyle\pm\,1.38}$ & $77.24\,{\scriptstyle\pm\,1.74}$ & ${+}7.36\,{\scriptstyle\pm\,1.88}$ \\
\midrule
Later-expert cost$\downarrow$ & -- & $5.93\,{\scriptstyle\pm\,0.32}$ & $9.01\,{\scriptstyle\pm\,1.15}$ & ${-}3.08\,{\scriptstyle\pm\,0.98}$ & -- & $10.02\,{\scriptstyle\pm\,0.86}$ & $18.31\,{\scriptstyle\pm\,1.55}$ & ${-}8.29\,{\scriptstyle\pm\,1.66}$ \\
\bottomrule
\end{tabular}
\end{adjustbox}
\end{table*}